\documentclass[11pt,a4paper]{article}
\usepackage{times,latexsym}
\usepackage{url}

\usepackage[T1]{fontenc}

\usepackage{natbib}
\usepackage{color,soul}
\usepackage{booktabs}
\usepackage{pifont}
\newcommand{\cmark}{\ding{51}}%
\newcommand{\xmark}{\ding{55}}%
\usepackage{tablefootnote}

\usepackage{comment}
\usepackage{enumitem}
\usepackage[tableposition=bottom]{caption}
\usepackage{graphicx}

\usepackage[acronym]{glossaries}
\usepackage{glossary-longbooktabs}

\newacronym{rnn}{RNN}{recurrent neural network}
\newacronym{seq2seq}{seq2seq}{sequence-to-sequence}
\newacronym{lstm}{LSTM}{long short-term memory}
\newacronym{gru}{GRU}{gated recurrent unit}
\newacronym{cnn}{CNN}{convolutional neural network}
\newacronym{relu}{ReLU}{rectified linear unit}
\newacronym{ctc}{CTC}{connectionist temporal classification}
\newacronym{recnn}{RecNN}{recursive neural network}
\newacronym{gan}{GAN}{generative adversarial network}

\newacronym{sgd}{SGD}{stochastic gradient descent}
\newacronym{em}{EM}{expectation maximization}
\newacronym{lrp}{LRP}{layer-wise relevance propagation}
\newacronym{svd}{SVD}{singular value decomposition}
\newacronym{lda}{LDA}{latent Dirichlet allocation}
\newacronym{svm}{SVM}{support vector machine}

\newglossaryentry{softmax}{name=Softmax, description={Normalization layer used in classification tasks}}
\newglossaryentry{attention}{name=attention, description={Mechanism for aligning source and target inputs in \acrfull{seq2seq} models}}
\newglossaryentry{dropout}{name=dropout, description={Method for regularizing neural networks}}
\newglossaryentry{highway}{name=highway, description={Type of connection in neural networks}}
\newglossaryentry{feedforward}{name=feed-forward, description={}}
\newglossaryentry{adam}{name=Adam, description={An adaptive optimization method}}

\newacronym[longplural={parts-of-speech}]{pos}{POS}{part-of-speech}
\newacronym{oov}{OOV}{out-of-vocabulary}
\newacronym{sem}{SEM}{semantic}

\newglossaryentry{lemma}{name=lemma, description={A dictionary item}}
\newglossaryentry{morpheme}{name=morpheme, description={A meaningful morphological unit}}
\newglossaryentry{phoneme}{name=phoneme, description={An abstract speech class that carries meaning in a specific language}}
\newglossaryentry{phone}{name=phone, description={A speech sound}}

\newacronym{pmb}{PMB}{Groningen Parallel Meaning Bank}
\newacronym{nlp}{NLP}{natural language processing}
\newacronym{pbmt}{PBMT}{phrase-based machine translation}
\newacronym{nmt}{NMT}{neural machine translation}
\newacronym{lsa}{LSA}{latent semantic analysis}
\newacronym{rte}{RTE}{recognizing textual entailment}
\newacronym{nli}{NLI}{natural language inference}

\newacronym{asr}{ASR}{automatic speech recognition}
\newacronym{mfcc}{MFCC}{Mel-frequency cepstral coefficient}
\newacronym{wer}{WER}{word error rate}
\newacronym{fft}{FFT}{fast Fourier transform}
\newacronym{lpc}{LPC}{linear predictive coding}
\newacronym{dtw}{DTW}{dynamic time warping}
\newacronym{hmm}{HMM}{hidden Markov model}
\newacronym{gmm}{GMM}{Gaussian mixture model}
\newacronym{plp}{PLP}{perceptual linear prediction}

\newglossaryentry{deepspeech2}{name=DeepSpeech2, description={End-to-end \gls{asr} model based on \gls{ctc}}}
\newglossaryentry{hamming}{name=Hamming, description={A window function often used in speech processing}} 
\newglossaryentry{formant}{name=formant, description={A natural frequency or resonance of a speech signal}}
\newglossaryentry{cepstrum}{name=cepstral analysis, description={A common procedure for de-convolution of a speech signal}}

\newglossaryentry{f1}{name=F$_1$, description={An evaluation metric defined as the harmonic mean of precision and recall}}

\newacronym{ai}{AI}{Artificial Intelligence}

\usepackage[acceptedWithA]{tacl2018v2}

\usepackage{xspace,mfirstuc,tabulary}

\newif\iftaclinstructions
\taclinstructionsfalse 
\iftaclinstructions
\renewcommand{\confidential}{}
\renewcommand{\anonsubtext}{(No author info supplied here, for consistency with
TACL-submission anonymization requirements)}
\newcommand{\instr}
\fi

\iftaclpubformat 

\else

\fi

\title{Analysis Methods in Neural Language Processing: A Survey}

\author{Yonatan Belinkov$^{12}$ \and James Glass$^1$ \\\\
  $^1$MIT Computer Science and Artificial Intelligence Laboratory \\ 
  $^2$Harvard School of Engineering and Applied Sciences \\
  Cambridge, MA, USA \\
   {\tt \{belinkov, glass\}@mit.edu } 
  }

\date{}

\begin{document}
\maketitle
\begin{abstract}
 The field of \acrlong{nlp} has seen impressive progress in recent years, with neural network models replacing many of the traditional systems. A plethora of new models have been proposed, many of which are thought to be opaque compared to their feature-rich counterparts. 
 This has led researchers to analyze, interpret, and evaluate neural networks in novel and more fine-grained ways. 
 In this survey paper, we review analysis methods in neural language processing, categorize them according to prominent research trends, highlight existing limitations, and point to potential directions for future work. 
 
\end{abstract}

\section{Introduction}

The rise of deep learning has transformed the field of \gls{nlp} in recent years. Models based on neural networks have obtained impressive improvements in various tasks, including language modeling~\cite{mikolov2010recurrent,jozefowicz2016exploring}, syntactic parsing~\cite{Q16-1023},  machine translation (MT)~\cite{bahdanau2014neural,sutskever2014sequence}, and many other tasks; see \citet{goldberg2017neural} for example success stories.  

This progress has been accompanied by a myriad of new neural network architectures. In many cases, traditional feature-rich systems are being replaced by end-to-end neural networks that aim to map input text to some output prediction. As end-to-end systems are gaining prevalence, one may point to two trends. First, some push back against the abandonment of linguistic knowledge and call for incorporating it inside the networks in different ways.\footnote{See, for instance, Noah Smith's invited talk at ACL 2017: \url{vimeo.com/234958746}. See also a recent debate on this matter  by Chris Manning and Yann LeCun: \url{www.youtube.com/watch?v=fKk9KhGRBdI}. (Videos accessed on December 11, 2018.)} 
Others strive to better understand how neural language processing models work. This theme of analyzing neural networks has connections to the broader work on interpretability in machine learning, along with specific characteristics of the \gls{nlp} field.

Why should we analyze our neural \gls{nlp} models? 
To some extent, this question falls into the larger question of interpretability in machine learning, which has been the subject of much debate in recent years.\footnote{See, for example, the NIPS 2017 debate: \url{www.youtube.com/watch?v=2hW05ZfsUUo}. (Accessed on December 11, 2018.)}
Arguments in favor of interpretability in machine learning usually mention goals 
like accountability, trust,  fairness, safety, and reliability~\cite{DoshiKim2017Interpretability,lipton2016mythos}. 
Arguments against typically stress performance as the most important desideratum. All these arguments naturally apply to machine learning applications in \gls{nlp}. 

In the context of \gls{nlp}, 
this question
needs to be understood in light of earlier \gls{nlp} work, often referred to as feature-rich or feature-engineered systems. In some of these systems, features are more easily understood by humans -- they can be morphological properties, lexical classes, syntactic categories, semantic relations, etc. In theory, one could observe the importance assigned by statistical \gls{nlp} models to such features in order to gain a better understanding of the model.\footnote{Nevertheless, one could question how feasible such an analysis is; 
consider for example interpreting support vectors in high-dimensional \glspl{svm}.} 
In contrast, it is more difficult to understand what happens in an end-to-end neural network model that takes input (say, word embeddings) and generates an output (say, a sentence classification). Much of the analysis work thus aims to understand how linguistic concepts that were common as features in \gls{nlp} systems are captured in neural networks. 

As the analysis of neural networks for language is becoming more and more prevalent, neural networks in various \gls{nlp} tasks are being analyzed; different network architectures and components are being compared; and a variety of new analysis methods are being developed. 
This survey aims to review and summarize this body of work, highlight current trends, and point to existing lacunae. It organizes the literature into several themes. Section~\ref{sec:ling} reviews work that targets a fundamental question: what kind of linguistic information is captured in neural networks? We also point to limitations in current methods for answering this question. 
Section~\ref{sec:viz} discusses visualization methods, and emphasizes the difficulty in evaluating visualization work. 
In Section~\ref{sec:challenge} we discuss the compilation of challenge sets, or test suites, for fine-grained evaluation, a methodology that has old roots in \gls{nlp}.  
Section~\ref{sec:adversarial} deals with the generation and use of adversarial examples to probe weaknesses of neural networks. 
We point to unique characteristics of dealing with text as a discrete input and how different studies handle them. 
Section~\ref{sec:explain} summarizes work on explaining model predictions, an important goal of interpretability research. This is a relatively under-explored area, and we call for more work in this direction.
Section~\ref{sec:other} mentions a few other methods that do not fall neatly into one of the above themes. 
In the conclusion, we summarize the main gaps and potential research directions for the field. 

The paper is accompanied by online supplementary materials that contain detailed references for studies corresponding to Sections~\ref{sec:ling}, \ref{sec:challenge}, and \ref{sec:adversarial} (Tables~\ref{tab:ling-info}, \ref{tab:challenge-sets}, and \ref{tab:adversarial}, 
respectively), available at \url{http://boknilev.github.io/nlp-analysis-methods}. 

\medskip 

Before proceeding, we briefly mention some earlier work of a similar spirit.

\paragraph{A historical note}
Reviewing the vast literature on neural networks for language is beyond our scope.\footnote{For instance, a neural network that learns distributed representations of words was developed already in \citet{Miikkulainen:1991}. 
See \citet[chapter 12.4]{Goodfellow-et-al-2016} for references to other important milestones. }
However, we mention here a few representative studies that focused on analyzing such networks, in order to illustrate how recent trends have roots that go back to before the recent deep learning revival.   

\citet{Rumelhart:1986:LPT:21935.42475} built a feed-forward neural network for learning the English past tense and analyzed its performance on a variety of examples and conditions. They were especially concerned with the performance over the course of training, as their goal was to model the past form acquisition in children. They also analyzed a scaled-down version having 8 input units and 8 output units, which allowed them to describe it exhaustively and examine how certain rules manifest in network weights. 

In his seminal work on \glspl{rnn}, Elman trained networks on synthetic sentences in a language prediction task~\cite{elman1989representation,elman1990finding,elman1991distributed}. Through extensive analyses, he showed how networks discover the notion of a word when predicting characters; capture syntactic structures like number agreement; and acquire word representations that reflect lexical and syntactic categories. Similar analyses were later applied to other networks and tasks~\cite{doi:10.1080/09540099008915660,doi:10.1080/09540090010014070,POLLACK199077,doi:10.1080/10489223.2013.796950}. 

While Elman's work was limited in some ways, such as evaluating generalization or various linguistic phenomena---as Elman himself recognized~\cite{elman1989representation}---it introduced methods that are still relevant today: 
from visualizing network activations in time, through clustering words by hidden state activations, to projecting representations to dimensions that emerge as capturing properties like sentence number or verb valency. The sections on visualization (Section~\ref{sec:viz}) and identifying linguistic information (Section~\ref{sec:ling}) contain many examples for these kinds of analysis.

\vspace{-3pt}
\section{What linguistic information is captured in neural networks} \label{sec:ling}
\vspace{-2pt}

Neural network models in \gls{nlp} are typically trained in an end-to-end manner on input-output pairs, without explicitly encoding linguistic features. Thus a primary questions is the following: what linguistic information is captured in neural networks? 
When examining answers to this question, it is convenient to consider three dimensions: 
which methods are used for conducting the analysis, 
 what kind of linguistic information is sought, and which objects in the neural network are being investigated.
 Table~\ref{tab:ling-info} 
 (in the supplementary materials) categorizes relevant analysis work according to these criteria. 
 In the next sub-sections, we discuss trends in analysis work along these lines, followed by a discussion of limitations of current approaches.

\subsection{Methods}
The most common approach for associating neural network components with linguistic properties is to predict such properties from activations of the neural network.
Typically, in this approach a neural network model is trained on some task (say, MT) and its weights are frozen. Then, the trained model is used for generating feature representations for another task  by running it on a corpus with linguistic annotations and recording the representations (say, hidden state activations). Another classifier is then used for predicting the property of interest (say, \gls{pos} tags). The performance of this classifier is used for evaluating the quality of the generated representations, and by proxy that of the original model. 
This kind of approach has been used in numerous papers in recent years; see Table~\ref{tab:ling-info} 
for references.\footnote{A similar method has been used to analyze hierarchical structure in neural networks trained on arithmetic expressions~\cite{veldhoen2016diagnostic,hupkes2017visualisation}.} 
It is referred to by various names, including ``auxiliary prediction tasks''~\cite{adi2016fine}, ``diagnostic classifiers''~\cite{veldhoen2016diagnostic}, and ``probing tasks''~\cite{conneau2018you}.

As an example of this approach, let us walk through an application to analyzing syntax in \gls{nmt} by \citet{shi-padhi-knight:2016:EMNLP2016}. In this work, two \gls{nmt} models were trained on standard parallel data -- English$\rightarrow$French and English$\rightarrow$German.  
The trained models (specifically, the encoders) were run on an annotated corpus and their hidden states were used for training a logistic regression classifier that predicts different syntactic properties. The authors concluded that the \gls{nmt} encoders learn significant syntactic information at both word-level and sentence-level. They also compared representations at different encoding layers and found that ``local features are somehow preserved in the lower layer whereas more global, abstract information tends to be stored in the upper layer.'' These results demonstrate the kind of insights that the classification analysis may lead to, especially when comparing different models or model components. 

Other methods for finding correspondences between parts of the neural network and certain properties include counting how often attention weights agree with a linguistic property like anaphora resolution~\cite{P18-1117} or  directly computing correlations between neural network activations and some property, for example, correlating \gls{rnn} state activations with depth in a syntactic tree~\cite{qian-qiu-huang:2016:EMNLP2016} or with \gls{mfcc} acoustic features~\cite{wu2016investigating}. Such correspondence  may also be computed indirectly. 
For instance, \citet{K17-1037} defined an ABX discrimination task to evaluate how a neural model of speech (grounded in vision) encoded phonology. Given phoneme representations from different layers in their model, and three phonemes, A, B, and X, they compared whether the model representation for X is closer to A or B. This discrimination task enabled them to draw conclusions about which layers encoder phonology better, observing that lower layers generally encode more phonological information.

\subsection{Linguistic phenomena}
Different kinds of linguistic information have been analyzed, ranging from basic properties like sentence length, word position, word presence, or simple word order, 
to morphological, syntactic, and semantic information. 
Phonetic/phonemic information, 
speaker information, 
and style and accent information 
have been studied in neural network models for speech, or in joint audio-visual models. 
See Table~\ref{tab:ling-info} 
for references. 

While it is difficult to synthesize a holistic picture from this diverse body of work, it appears that neural networks are able to learn a substantial amount of information on various linguistic phenomena. 
These models are especially successful at capturing frequent properties, while some rare properties are more difficult to learn. 
\citet{linzen2016assessing}, for instance, found that \gls{lstm} language models are able to capture subject-verb agreement in many common cases, while direct supervision is required for solving harder cases.

Another theme that emerges in several studies is the hierarchical nature of the learned representations. We have already mentioned such findings regarding \gls{nmt} \cite{shi-padhi-knight:2016:EMNLP2016} and a visually grounded speech model \cite{K17-1037}. Hierarchical representations of syntax were also reported to emerge in other \gls{rnn} models \cite{P18-2003}. 

Finally, a  couple of papers discovered that models trained with latent trees perform better on \gls{nli}~\cite{Q18-1019,W18-2903} than ones trained with linguistically-annotated trees. 
Moreover, the trees in these models do not resemble syntactic trees corresponding to known linguistic theories, which casts doubts on the importance of syntax-learning in the underlying neural network.\footnote{Others found that even simple binary trees may work well in MT~\cite{wang18emnlptrdec} and sentence classification~\cite{D15-1092}.}

\subsection{Neural network components}
In terms of the object of study, various neural neural network components were investigated, including  word embeddings, 
\gls{rnn} hidden states or gate activations, 
sentence embeddings, 
and \gls{attention} weights in \gls{seq2seq} models. 
 Generally less work has analyzed \glspl{cnn} in \gls{nlp}, but see \citet{W18-5408} for a recent exception. 
In speech processing, researchers have analyzed  layers in deep neural networks for speech recognition 
and different speaker embeddings. 
Some analysis has also been devoted to  joint language-vision 
or audio-vision models, 
or to similarities between word embeddings and convolutional image representations. 
Table~\ref{tab:ling-info} 
provides detailed references.

\subsection{Limitations}

The classification approach may find that a certain amount of linguistic information is captured in the neural network. However, this does not necessarily mean that the information is used by the network. 
For example, 
\newcite{vanmassenhoveinvestigating} investigated aspect in \gls{nmt} (and in phrase-based statistical MT). They trained a classifier on \gls{nmt} sentence encoding vectors and found that they can accurately predict tense about $90$\% of the time. However, when evaluating the output translations, they found them to have the correct tense only $79$\% of the time. They interpreted this result to mean that ``part of the aspectual information is lost during decoding''. 
Relatedly, \citet{P18-1126} compared the performance of various \gls{nmt} models in terms of translation quality (BLEU) and representation quality (classification tasks). They found a negative correlation between the two, suggesting  that high-quality systems may not be learning certain sentence meanings.  
In contrast, \citet{K18-1028} showed that word embeddings contain divergent linguistic information, which can be uncovered by applying a linear transformation on the learned embeddings. Their results suggest an alternative explanation, showing that ``embedding models are able to encode divergent linguistic information but have limits on how this information is surfaced.'' 

From a methodological point of view, most of the relevant analysis work is concerned with \emph{correlation}: how correlated are neural network components with linguistic properties? What may be lacking is a measure of \emph{causation}: how does the encoding of linguistic properties affect the system output. 
 \newcite{giulianelli2018under} make some headway on this question. They predicted number agreement from \gls{rnn} hidden states and gates at different time steps. They then intervened in how the model processes the sentence by changing a hidden activation based on the difference between the prediction and the correct label. This improved agreement prediction accuracy, and the effect persisted over the course of the sentence, indicating that this information has an effect on the model. However, they did not report the effect on  overall model quality, for example by measuring perplexity. 
 Methods from causal inference may shed new light on some of these questions. 

Finally, the predictor for the auxiliary task is usually a simple classifier, such as logistic regression. 
A few studies compared different classifiers and found that deeper classifiers lead to overall better results, but do not alter the respective trends when comparing different models or components~\cite{qian-qiu-huang:2016:P16-11,belinkov:2018:phdthesis}.
Interestingly, \newcite{conneau2018you} found that tasks requiring more nuanced linguistic knowledge (e.g., tree depth, coordination inversion) gain the most from using a deeper classifier. 
However, the approach is usually taken for granted; given its prevalence, it appears that better theoretical or empirical foundations are in place.


\section{Visualization} \label{sec:viz}

\begin{figure}[t]
\centering
\includegraphics[width=\linewidth]{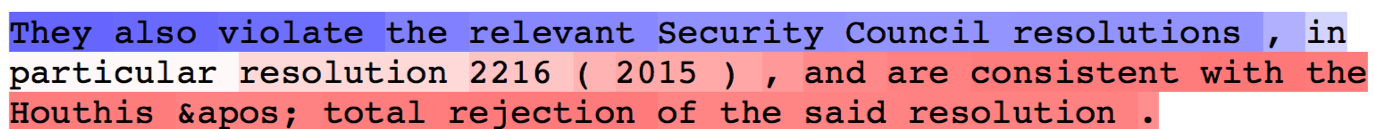}
\caption{A heatmap visualizing neuron activations. In this case, the activations capture position in the sentence.}
\label{fig:viz-position}
\end{figure}

Visualization is a valuable tool for analyzing neural networks in the language domain and beyond. 
Early work visualized hidden unit activations in \glspl{rnn} trained on an artificial language modeling task, and observed how they  correspond to certain grammatical relations such as agreement~\cite{elman1991distributed}. Much recent work has focused on visualizing activations on specific examples in modern neural networks for language~\cite{karpathy2015visualizing,kadar2016representation,qian-qiu-huang:2016:EMNLP2016,W18-3024} and speech~\cite{wu2016investigating,nagamine2015exploring,wang2017gate}. 
 Figure~\ref{fig:viz-position} shows an example visualization of a neuron that captures position of words in a sentence. The heatmap uses blue and red colors for negative and positive activation values, respectively, enabling the user to quickly grasp the function of this neuron.
 
\begin{figure}[t]
\centering
\includegraphics[width=0.8\linewidth]{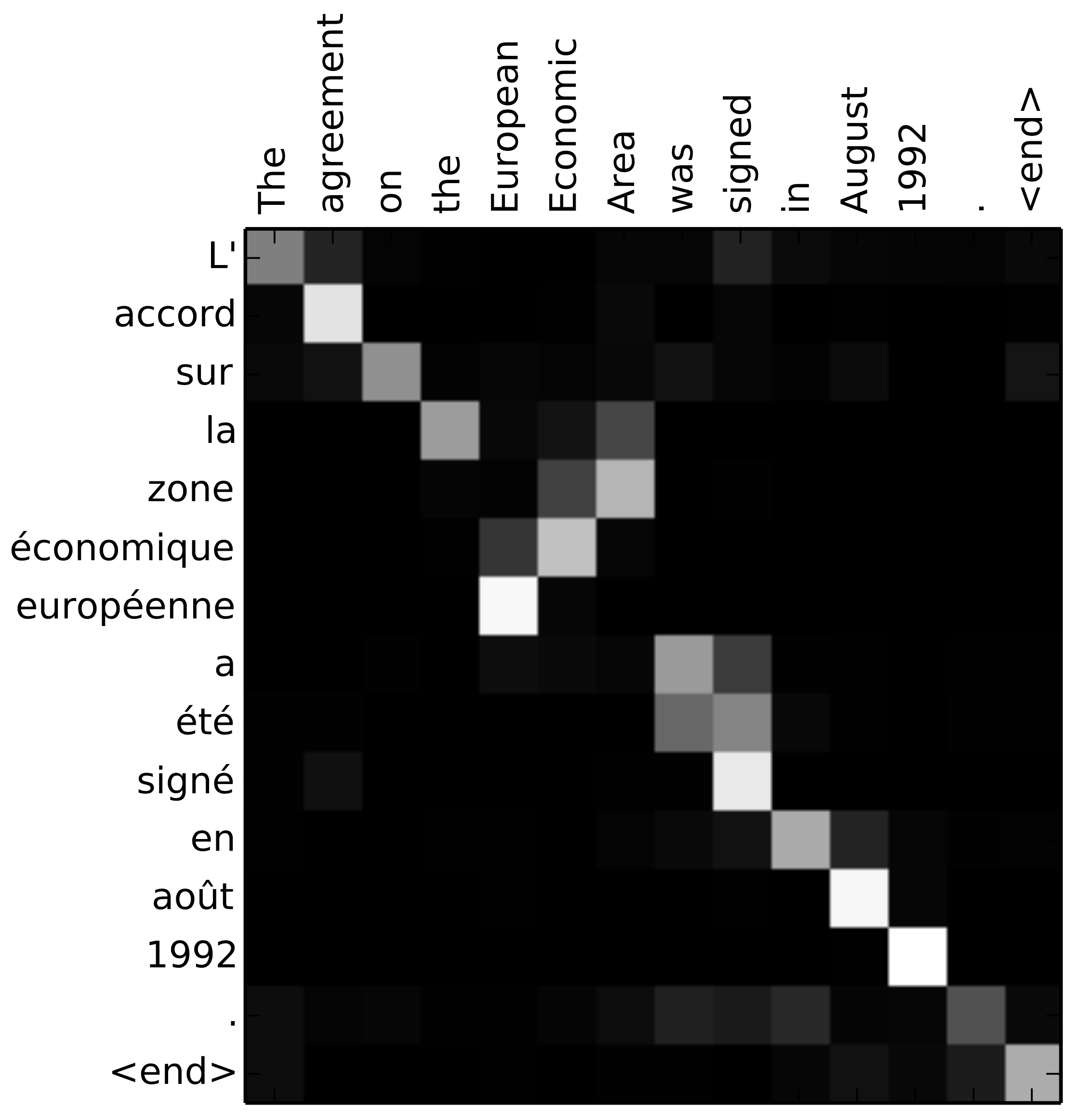}
\caption{A visualization of attention weights, showing soft alignment between source and target sentences in an \gls{nmt} model. Reproduced from \citet{bahdanau2014neural}, with permission.}
\label{fig:viz-attention}
\end{figure}

 The \gls{attention} mechanism that originated in work on \gls{nmt}~\cite{bahdanau2014neural} also lends itself to a natural visualization. The alignments obtained via different attention mechanisms  have produced visualizations ranging from tasks like 
 \gls{nli} \cite{rocktaschel2016reasoning,Q16-1019},  summarization~\cite{D15-1044}, MT post-editing~\cite{unanue2018shared}, and morphological inflection \cite{P17-1183}, to matching users on social media \cite{tay2018couplenet}. 
 Figure~\ref{fig:viz-attention} reproduces a visualization of attention alignments from the original work by  \citeauthor{bahdanau2014neural}. Here grayscale values correspond to the weight of the attention between words in an English source sentence (columns) and its French translation (rows). As \citeauthor{bahdanau2014neural} explain, this visualization demonstrates that the \gls{nmt} model learned a soft alignment between source and target words. Some aspects of word order may also be noticed, as in the reordering of noun and adjective when translating the phrase ``European Economic Area''.

Another line of work computes various saliency measures to attribute predictions to input features. The important or salient features can then be visualized in selected examples~\cite{N16-1082,D16-1216,pmlr-v70-sundararajan17a,10.1371/journal.pone.0181142,W17-5221,P17-1106,james2018beyond,P18-1176,MONTAVON20181,godin2018explaining}. 
Saliency can also be computed with respect to intermediate values, rather than input features~\cite{ghaeini2018interpreting}.\footnote{Generally, many of the visualization methods are adapted from the vision domain, where they have been extremely popular; see \citet{Zhang2018} for a survey.}

An instructive visualization technique is to cluster neural network activations and compare them to some linguistic property. Early work clustered \gls{rnn} activations, showing that they organize in lexical categories~\cite{elman1989representation,elman1990finding}. Similar techniques have been followed by others. Recent examples include clustering of sentence embeddings in an \gls{rnn} encoder trained in a multi-task learning scenario~\cite{brunner2018natural}, and \gls{phoneme} clusters in a joint audio-visual \gls{rnn} model~\cite{K17-1037}.

A few online tools for visualizing neural networks have recently become available. \texttt{LSTMVis}~\cite{strobelt2018lstmvis} visualizes \gls{rnn} activations, focusing on tracing hidden state dynamics.\footnote{\texttt{RNNVis}~\cite{ming2017understanding} is a similar tool, but its online demo does not seem to be available at the time of writing.} 
\texttt{Seq2Seq-Vis} \cite{strobelt2018seq2seq}  visualizes different modules in \gls{attention}-based \gls{seq2seq} models, with the goal of examining model decisions and testing alternative decisions. 
Another tool focused on comparing attention alignments was proposed by \citet{rikters2018debugging}. 
It also provides translation confidence scores based on the distribution of attention weights.
\texttt{NeuroX}~\cite{dalvi:2019:AAAI:demo} is a tool for finding and analyzing individual neurons, focusing on machine translation.

\paragraph{Evaluation}
As in much work on interpretability, evaluating visualization quality is difficult and often limited to qualitative examples.  A few notable exceptions report human evaluations of  visualization quality. 
\newcite{singh2018hierarchical} showed humans hierarchical clusterings 
of input words generated by two interpretation methods,  and asked them to evaluate which method is more accurate, or in which method they trust more. 
Others reported human evaluations for attention visualization  in conversation modeling \cite{freeman2018paying} and medical code prediction tasks \cite{N18-1100}.

 The availability of open-source tools of the sort described above will hopefully encourage users to utilize visualization in their regular research and development cycle. 
 However, it remains to be seen how useful visualizations turn out to be.


\section{Challenge sets} \label{sec:challenge}

The majority of benchmark datasets in \gls{nlp} are drawn from text corpora, reflecting a natural frequency distribution of language phenomena. While useful in practice for evaluating system performance in the average case, such datasets may fail to capture a wide range of phenomena. 
An alternative evaluation framework consists of challenge sets, also known as test suites, which have been used in \gls{nlp} for a long time~\cite{C96-2120}, especially for evaluating  MT systems~\cite{C90-2037,isahara1995jeida,koh2001test}. 
\citet{C96-2120} noted several key properties of test suites: systematicity, control over data, inclusion of negative data, and exhaustivity. They contrasted such datasets with test corpora, ``whose main advantage is that they reflect naturally occurring data.'' This idea underlines much of the work on challenge sets and is echoed in more recent work~\cite{wang2018glue}. 
For instance, \newcite{cooper1996using} constructed a semantic test suite that targets phenomena as diverse as quantifiers, plurals, anaphora, ellipsis, adjectival properties, and so on.

After a hiatus of a couple of decades,\footnote{One could speculate that their decrease in popularity can be attributed to the rise of large-scale quantitative evaluation of statistical \gls{nlp} systems. }  
challenge sets have recently gained renewed popularity in the \gls{nlp} community. 
In this section, we include datasets used for evaluating neural network models that diverge from the common average-case evaluation. Many of them share some of the properties noted by \citet{C96-2120}, although negative examples (ill-formed data) are typically less utilized. 
The challenge 
 datasets can be categorized along the following criteria: the task they seek to evaluate, the linguistic phenomena they aim to study, the language(s) they target, their size, their method of construction, and how performance is evaluated.\footnote{Another typology of evaluation protocols was put forth by \newcite{W17-4705}.  
Their criteria are partially overlapping with ours, although they did not provide a comprehensive categorization as the one compiled here.}  Table~\ref{tab:challenge-sets} 
(in the supplementary materials) categorizes many recent challenge sets along these criteria. Below we discuss common trends along these lines.

\subsection{Task}
By far, the most targeted tasks in challenge sets are \gls{nli} and MT. 
This can partly be explained by the popularity of these tasks and the prevalence of neural models proposed for solving them. 
Perhaps more importantly, tasks like \gls{nli} and MT arguably require inferences at various linguistic levels, making the challenge set evaluation especially attractive. 
Still, other high-level tasks like reading comprehension or question answering have not received as much attention, and may also benefit from the careful construction of challenge sets. 

A significant body of work aims to evaluate the quality of embedding models  
by correlating the similarity they induce on word or sentence pairs with human similarity judgments. Datasets containing such similarity scores are often used to evaluate 
word embeddings~\cite[inter alia]{finkelstein2002placing,P12-1015,J15-4004} or sentence embeddings; see the many shared tasks on semantic textual similarity in SemEval~\cite[and previous editions]{S17-2001}. 
Many of these datasets evaluate similarity at a coarse-grained level, but some provide a more fine-grained evaluation of similarity or relatedness. 
For example,  
some datasets are dedicated for specific word classes such as verbs~\cite{D16-1235} or rare words~\cite{W13-3512}, or for evaluating compositional knowledge in sentence embeddings~\cite{S14-2001}. Multilingual and cross-lingual versions have also been collected~\cite{leviant2015separated,S17-2001}.  Although these datasets are widely used, this kind of evaluation has been criticized for its subjectivity and questionable correlation with downstream performance~\cite{Faruqui:repeval:16}. 

\subsection{Linguistic phenomena}
One of the primary goals of challenge sets is to evaluate models on their ability to 
handle specific linguistic phenomena. While earlier studies emphasized exhaustivity~\cite{cooper1996using,C96-2120},  recent ones tend to focus on a few properties of interest.  
For example, \newcite{E17-2060} introduced a challenge set for MT evaluation focusing on $5$ properties: subject-verb agreement, noun phrase agreement, verb-particle constructions, polarity, and transliteration. 
 Slightly more elaborated is an MT challenge set for morphology, including $14$ morphological properties~\cite{W17-4705}.
See Table~\ref{tab:challenge-sets} 
for references to datasets targeting other phenomena. 

Other challenge sets cover a more diverse range of linguistic properties, in the spirit of some of the earlier work. 
For instance, 
extending the categories in \citet{cooper1996using}, the GLUE analysis set for \gls{nli} covers more than $30$ phenomena in four coarse categories (lexical semantics, predicate-argument structure, logic, and knowledge). 
In MT evaluation, \newcite{burchardt2017linguistic} reported results using a large test suite covering $120$ phenomena, partly based on \citet{C96-2120}.\footnote{Their dataset does not seem to be available yet, but more details are promised to appear in a future publication.}
\newcite{D17-1263} and \newcite{isabelle2018challenge} prepared challenge sets for MT evaluation covering fine-grained phenomena at morpho-syntactic, syntactic, and lexical levels.

Generally, datasets that are constructed programmatically tend to cover less fine-grained linguistic properties, while manually constructed datasets represent more diverse phenomena.

\subsection{Languages}
As unfortunately usual in much \gls{nlp} work, especially neural \gls{nlp}, the vast majority of challenge sets are  in English. This situation is slightly better in MT evaluation, where naturally all datasets feature other languages (see Table~\ref{tab:challenge-sets}). 
A notable exception is the work by \newcite{gulordava2018colorless}, who constructed examples for evaluating number agreement in language modeling in English, Russian, Hebrew, and Italian. 
Clearly, there is room for more challenge sets in non-English languages. However, perhaps more pressing is the need for large-scale non-English datasets (besides MT) to develop neural models for popular \gls{nlp} tasks.  

\subsection{Scale}
The size of proposed challenge sets varies greatly (Table~\ref{tab:challenge-sets}). 
As expected, datasets constructed by hand are smaller, with typical sizes in the hundreds. Automatically-built datasets are much larger, ranging from several thousands to  close to a hundred thousand \cite{E17-2060}, or even more than one million examples~\cite{linzen2016assessing}. In the latter case, the authors argue that such a large test set is needed for obtaining a sufficient representation of rare cases. 
A few manually-constructed datasets contain a fairly large number of examples, up to $10$K 
\cite{burchardt2017linguistic}.  

\subsection{Construction method}
Challenge sets are usually created either programmatically or manually, by hand-crafting specific examples. Often, semi-automatic methods are used to compile an initial list of examples that is manually verified by annotators. The specific method also affects the kind of language use and how natural or artificial/synthetic the examples are. 
We describe here some trends in dataset construction methods in the hope that they may be useful for researchers contemplating new datasets.

Several datasets were constructed by modifying or extracting examples from existing datasets. For instance,
\newcite{N18-1179} and \newcite{P18-2103} extracted examples from SNLI~\cite{D15-1075} and replaced specific words such as hypernyms, synonyms, and antonyms, followed by manual verification. 
\newcite{linzen2016assessing}, on the other hand, extracted examples of subject-verb agreement from raw texts using heuristics, resulting in a large-scale dataset. 
\newcite{gulordava2018colorless} extended this to other agreement phenomena, but they relied on syntactic information available in treebanks, resulting in a smaller dataset. 

Several challenge sets utilize existing test suites, either as a direct source of examples~\cite{burchardt2017linguistic} or for searching similar naturally occurring examples~\cite{wang2018glue}.\footnote{\citet{wang2018glue} also verified that their examples do 
not contain annotation artifacts, a potential problem noted in recent studies~\cite{N18-2017,S18-2023}.}

\newcite{E17-2060} introduced a method for evaluating \gls{nmt} systems via \textit{contrastive translation pairs}, where the system is asked to estimate the probability of two candidate translations that are designed to reflect specific linguistic properties. 
\citeauthor{E17-2060} generated such pairs programmatically by applying simple heuristics, 
such as changing gender and number to induce agreement errors, resulting in a large-scale challenge set of close to $100$K examples.  
This framework was extended to evaluate other properties, but often requiring more sophisticated generation methods like using morphological analyzers/generators~\cite{W17-4705} or more manual involvement in generation~\cite{N18-1118} or verification~\cite{W17-4702}.

Finally, a few of studies define templates that capture certain linguistic 
properties and instantiate them with word lists~\cite{dasgupta2018evaluating,N18-2002,N18-2003}. 
Template-based generation has the advantage of providing more control, for example for obtaining a specific vocabulary distribution, but this comes at the expense of how natural the examples are.

\subsection{Evaluation}
Systems are typically evaluated by their performance on the challenge set examples, either with the same metric used for evaluating the system in the first place, or via a proxy, as in the  contrastive pairs evaluation of \citet{E17-2060}. 
Automatic evaluation metrics are cheap to obtain and can be calculated on a large scale. However, they may miss certain aspects. Thus a few studies report human evaluation on their challenge sets, such as in MT \cite{D17-1263,burchardt2017linguistic}. 

We note here also that judging the quality of a model by its performance on a challenge set can be tricky. Some authors emphasize their 
wish to test systems on extreme or difficult cases, ``beyond normal operational capacity'' \cite{naik2018stress}. However,  whether or not one should expect systems to perform well on specially chosen cases (as opposed to the average case) may depend 
on one's goals.
To put results in perspective, one may compare model performance to human performance on the same task~\cite{gulordava2018colorless}.


\section{Adversarial examples} \label{sec:adversarial}

Understanding a model requires also an understanding of its failures. 
Despite their success in many tasks, machine learning systems can also be very sensitive to malicious attacks or adversarial examples~\citep{szegedy2013intriguing,goodfellow2014explaining}. 
In the  vision domain, small changes to the input image can lead to misclassification, even if such changes are indistinguishable by humans.

The basic setup in work on adversarial examples can be described as follows.\footnote{The notation here follows \citet{yuan2017adversarial}.} 
Given a neural network model $f$ and an input example $x$, we seek to generate an adversarial example $x'$ that will have a minimal distance from $x$, while being assigned a different label by $f$: 
\begin{align*}
\min_{x'} &\null ||x - x'||  \\
\text{s.t.} &\null \hspace{10pt} f(x) = l, 
             f(x') = l', 
             l \neq l' 
\end{align*}
In the vision domain, $x$ can be the input image pixels, resulting in a fairly intuitive interpretation of this optimization problem: measuring the distance $||x-x'||$ is straightforward, and finding $x'$ can be done by computing gradients with respect to the input, since all quantities are continuous. 

In the text domain, the input is discrete (for example, a sequence of words), 
which poses two problems. First, it is not clear how to measure the distance between the original and adversarial examples, $x$ and $x'$, 
which are two discrete objects (say, two words or sentences). Second, minimizing this distance cannot be easily formulated as an optimization problem, as this requires computing gradients with respect to a discrete input.

In the following, we review methods for handling these difficulties according to several criteria:
the adversary's knowledge, 
the specificity of the attack, 
the linguistic unit being modified, 
and the task on which the attacked model was trained.\footnote{These criteria are partly taken from~\citet{yuan2017adversarial}, where a more elaborate taxonomy is laid out. At present, though, the work on adversarial examples in \gls{nlp} is more limited than 
in computer vision, so our criteria will suffice.}
Table~\ref{tab:adversarial} 
(in the supplementary materials) categorizes work on adversarial examples in \gls{nlp} according to these criteria. 

\subsection{Adversary's knowledge}

Adversarial examples can be generated using access to model parameters, also known as white-box attacks, or without such access, with black-box attacks~\citep{papernot2016transferability,Papernot:2017:PBA:3052973.3053009,8014906,liu2016delving}.

White-box attacks are difficult to adapt to the text world as they typically require computing gradients with respect to the input, which would be discrete in the text case. 
One option is to compute gradients with respect to the input word embeddings, and perturb the embeddings. Since this may result in a vector that does not correspond to any word, one could search for the closest word embedding in a given dictionary~\cite{papernot2016crafting}; 
\citet{cheng2018seq2sick} extended this idea to \gls{seq2seq} models. 
Others  computed gradients with respect to input word embeddings to identify and rank words to be modified~\cite{samanta2017towards,liang2017deep}. 
\citet{P18-2006} developed an alternative method by representing text edit operations in vector space (e.g., a binary vector specifying which characters in a word would be changed) and approximating the change in loss with the derivative along this vector. 

Given the difficulty in generating white-box adversarial examples for text, much research has been devoted to black-box examples. Often, the adversarial examples are inspired by text edits that are thought to be natural or commonly generated by humans, such as typos, misspellings, and so on~\cite{DBLP:conf/aaai/SakaguchiDPD17,heigold2017robust,belinkov:2018:ICLR}. 
\citet{gao2018black} defined scoring functions to identify tokens to modify. Their functions do not require access to model internals, but they do require the model prediction score. After identifying the important tokens, they modify characters with common edit operations. 

\citet{zhao2018generating} used \glspl{gan}~\cite{goodfellow2014generative} to minimize the distance between latent representations of input and adversarial examples, and performed perturbations in latent space. Since the latent representations do not need to come from the attacked model, this is a black-box attack.

Finally, \citet{alzantot2018generating} developed an interesting population-based genetic algorithm for crafting adversarial examples for text classification, by maintaining a population of modifications of the original sentence and evaluating fitness of modifications at each generation.  
 They do not require access to model parameters, but do use prediction scores. 
A similar idea was proposed by \citet{kuleshov2018adversarial}. 

\subsection{Attack specificity}
Adversarial attacks can be classified 
to targeted vs.\ non-targeted attacks~\cite{yuan2017adversarial}. A targeted attack specifies a specific false class,  $l'$, 
while a non-targeted attack only cares that the predicted class is wrong, $l' \neq l$. 
Targeted attacks are more difficult to generate, as they typically require knowledge of model parameters, i.e., they are white-box attacks. This might explain why the majority of adversarial examples in \gls{nlp} are non-targeted (see Table~\ref{tab:adversarial}). 
A few targeted attacks include \citet{liang2017deep}, which specified a desired class to fool a text classifier, and \citet{P18-1241}, which specified words or captions to generate in an image captioning model. 
Others targeted specific words to omit, replace, or include when attacking \gls{seq2seq} models~\cite{cheng2018seq2sick,C18-1055}.

Methods for generating targeted attacks in \gls{nlp} could possibly take more inspiration from adversarial attacks in other fields. For instance, in attacking malware detection systems, several studies developed targeted attacks in a black-box scenario~\cite{yuan2017adversarial}. 
A black-box targeted attack for MT was proposed by \citet{zhao2018generating}, who used \glspl{gan} to search for attacks on Google's MT system after mapping sentences into continuous space with adversarially regularized autoencoders~\cite{pmlr-v80-zhao18b}.

\subsection{Linguistic unit}
Most of the work on adversarial text examples involves modifications at the 
character- and/or word-level; see Table~\ref{tab:adversarial} 
for specific references. 
Other transformations include adding sentences or text chunks~\cite{jia-liang:2017:EMNLP2017} or generating paraphrases with desired syntactic structures~\cite{N18-1170}. In image captioning, \citet{P18-1241}  modified pixes in the input image to generate targeted attacks on the caption text. 

\subsection{Task}

Generally, most work on adversarial examples in \gls{nlp} concentrates on relatively high-level language understanding tasks, such as 
text classification (including sentiment analysis) and reading comprehension,
while work on text generation focuses mainly on MT. See Table~\ref{tab:adversarial} 
for references. 
There is relatively little work on adversarial examples for more low-level language processing 
tasks, although one can mention morphological tagging~\cite{heigold2017robust} and spelling correction~\cite{DBLP:conf/aaai/SakaguchiDPD17}. 

\subsection{Coherence \& perturbation measurement}
In adversarial image examples, it is fairly straightforward to measure the perturbation, either by measuring  distance in  pixel space, say $||x-x'||$ under some norm, 
or with alternative measures that are better correlated with human perception~\cite{rozsa2016adversarial}. It is also visually compelling to present an adversarial image with imperceptible difference from its source image. 
In the text domain, measuring distance is not as straightforward and even small changes to the text may be perceptible by humans. Thus, evaluation of attacks is fairly tricky. 
Some studies imposed constraints on adversarial examples to have a small number of edit operations~\cite{gao2018black}. Others ensured syntactic or semantic coherence in different ways, such as filtering replacements by word similarity or sentence similarity \cite{alzantot2018generating,kuleshov2018adversarial}, or by using synonyms and other word lists~\cite{samanta2017towards,yang2018greedy}.

Some 
reported whether a human can classify the adversarial example correctly~\cite{yang2018greedy}, but this does not indicate how perceptible the changes are. More informative human studies evaluate grammaticality or similarity of the adversarial examples to the original ones~\cite{zhao2018generating,alzantot2018generating}. Given the inherent difficulty in generating imperceptible changes in text, more such evaluations are needed.


\section{Explaining predictions} \label{sec:explain}

Explaining specific  
predictions is recognized as a desideratum in intereptability work~\cite{lipton2016mythos}, argued to increase the accountability of machine learning systems~\cite{doshi2017accountability}. 
However, explaining why a deep, 
highly non-linear neural network makes a certain prediction is not trivial. 
One solution is to ask the model to generate explanations along with its primary prediction~\cite{N07-1033,D16-1076},\footnote{Other  work considered learning textual-visual explanations from multi-modal annotations~\cite{park2018multimodal}.}  
but this approach requires manual annotations of explanations, which may be 
hard to collect.  

An alternative approach is to use parts of the input as explanations. 
 For example, \newcite{D16-1011} defined a generator that learns a 
 distribution over text fragments as candidate rationales for justifying predictions, evaluated on  sentiment analysis. \newcite{D17-1042} discovered input-output associations in a \acrlong{seq2seq} learning scenario,  by perturbing the input and finding the most relevant associations. 
\newcite{gupta2018lisa} inspected how information is accumulated in \glspl{rnn} towards a prediction, and associated peaks in prediction scores with important input segments. As these methods use input segments to explain predictions, they do not shed much light on the internal computations that take place in the network.

At present, despite the recognized importance for interpretability, our ability to explain predictions of neural networks in \gls{nlp} is still limited.


\section{Other methods} \label{sec:other}

We briefly mention here several analysis methods that do not fall neatly into the previous sections. 

A number of studies evaluated the effect of erasing or masking certain neural network components, such as word embedding dimensions, hidden units, or even full words~\cite{li2016understanding,FengRAWR2018,P18-1027,bau2018identifying}. 
For example, \citet{li2016understanding} erased specific dimensions in word embeddings or hidden states and computed the change in probability assigned to different labels. Their experiments revealed interesting differences between word embedding models, where in some models information is more focused in individual dimensions. They also found that information is more distributed in hidden layers than in the input layer, and erased entire words to find important words in a sentiment analysis task.

Several studies conducted behavioral experiments to interpret word embeddings by defining intrusion tasks, where humans need 
to identify an intruder word, 
chosen based on difference in word embedding dimensions~\cite{C12-1118,N15-1004,P15-1144}.\footnote{The methodology follows earlier work on evaluating the interpretability of probabilistic topic models with intrusion tasks \cite{NIPS2009_3700}.} In this kind of work, a word embedding model may be deemed more interpretable if humans are better able to identify the intruding words. 
Since the evaluation is costly for high-dimensional representations, alternative automatic metrics were considered~\cite{D17-1041,senel2018semantic}.

A long tradition in work on neural networks is to evaluate and analyze their ability to learn different formal languages~\cite{das1992learning,casey1996dynamics,gers2001lstm,boden2002learning,chalup2003incremental}. This trend continues today, with research into modern architectures and what formal languages they can learn~\cite{P18-2117,bernardy2018can,suzgun:2019:SCiL}, or the formal properties they possess~\cite{N18-1205}.  


\section{Conclusion} \label{sec:conclusion}

Analyzing neural networks has become a hot topic in \gls{nlp} research. This survey attempted to review and summarize as much of the current research as possible, while organizing it along several prominent themes. We have emphasized aspects in analysis that are specific to language -- namely, what linguistic information is captured in neural networks, which phenomena they are successful at capturing, and where they fail. Many of the analysis methods are general techniques from the larger machine learning community, such as visualization via saliency measures, or evaluation by adversarial examples. But even those sometimes require non-trivial adaptations to work with text input.   Some methods are more specific to the field, but may prove useful in other domains. Challenge sets or test suites are such a case. 

Throughout this survey, we have identified several limitations or gaps in current analysis work: 
\begin{itemize}
\item The use of auxiliary classification tasks for identifying which linguistic properties neural networks capture has become standard practice (Section~\ref{sec:ling}), while lacking both a theoretical foundation and a better empirical consideration of the link between the auxiliary tasks and the original task. 
\item Evaluation of analysis work is often limited or qualitative, especially in visualization techniques (Section~\ref{sec:viz}). Newer forms of evaluation are needed for determining the success of different methods. 
\item Relatively little work has been done on explaining predictions of neural network models, apart from providing visualizations (Section~\ref{sec:explain}). With the increasing public demand for explaining algorithmic choices in machine learning systems~\cite{DoshiKim2017Interpretability,doshi2017accountability}, there is pressing need for progress in this direction.  
\item Much of the analysis work is focused on the English language, especially in constructing challenge sets for various tasks (Section~\ref{sec:challenge}), with the exception of MT due to its inherent multilingual character.  Developing resources and evaluating methods on other languages is important as the field grows and matures. 
\item More challenge sets for evaluating other tasks besides \gls{nli} and MT are needed. 
\end{itemize}

Finally, as with any survey in a rapidly evolving field, this paper is likely to omit relevant recent work by the time of publication. 
While we intend to continue updating the online appendix with newer publications, we hope that our summarization of prominent analysis work and its categorization into several themes will be a useful guide for scholars interested in analyzing and understanding neural networks for \gls{nlp}.


\iftaclpubformat

\section*{Acknowledgments}

We would like to thank the anonymous reviewers and the TACL Action Editor for their very helpful comments. 
This work was supported by the Qatar Computing Research Institute. Y.B.\ is also supported by the Harvard Mind, Brain, Behavior Initiative. 
\else
\fi

\nocite{*}

\bibliography{tacl2018-camera-ready}

\begin{thebibliography}{193}
\expandafter\ifx\csname natexlab\endcsname\relax\def\natexlab#1{#1}\fi

\bibitem[{Adi et~al.(2017{\natexlab{a}})Adi, Kermany, Belinkov, Lavi, and
  Goldberg}]{adi2017analysis}
Yossi Adi, Einat Kermany, Yonatan Belinkov, Ofer Lavi, and Yoav Goldberg.
  2017{\natexlab{a}}.
\newblock Analysis of sentence embedding models using prediction tasks in
  natural language processing.
\newblock \emph{IBM Journal of Research and Development}, 61(4):3--9.

\bibitem[{Adi et~al.(2017{\natexlab{b}})Adi, Kermany, Belinkov, Lavi, and
  Goldberg}]{adi2016fine}
Yossi Adi, Einat Kermany, Yonatan Belinkov, Ofer Lavi, and Yoav Goldberg.
  2017{\natexlab{b}}.
\newblock {Fine-grained Analysis of Sentence Embeddings Using Auxiliary
  Prediction Tasks}.
\newblock In \emph{International Conference on Learning Representations
  (ICLR)}.

\bibitem[{Aharoni and Goldberg(2017)}]{P17-1183}
Roee Aharoni and Yoav Goldberg. 2017.
\newblock \href {https://doi.org/10.18653/v1/P17-1183} {{Morphological
  Inflection Generation with Hard Monotonic Attention}}.
\newblock In \emph{Proceedings of the 55th Annual Meeting of the Association
  for Computational Linguistics (Volume 1: Long Papers)}, pages 2004--2015.
  Association for Computational Linguistics.

\bibitem[{Ahmad et~al.(2018)Ahmad, Bai, Huang, Jiang, Peng, and
  Chang}]{ahmad2018multi}
Wasi~Uddin Ahmad, Xueying Bai, Zhechao Huang, Chao Jiang, Nanyun Peng, and
  Kai-Wei Chang. 2018.
\newblock {Multi-task Learning for Universal Sentence Embeddings: A Thorough
  Evaluation using Transfer and Auxiliary Tasks}.
\newblock \emph{arXiv preprint arXiv:1804.07911v2}.

\bibitem[{Alishahi et~al.(2017)Alishahi, Barking, and Chrupa{\l}a}]{K17-1037}
Afra Alishahi, Marie Barking, and Grzegorz Chrupa{\l}a. 2017.
\newblock \href {https://doi.org/10.18653/v1/K17-1037} {Encoding of phonology
  in a recurrent neural model of grounded speech}.
\newblock In \emph{Proceedings of the 21st Conference on Computational Natural
  Language Learning (CoNLL 2017)}, pages 368--378. Association for
  Computational Linguistics.

\bibitem[{Alvarez-Melis and Jaakkola(2017)}]{D17-1042}
David Alvarez-Melis and Tommi Jaakkola. 2017.
\newblock \href {http://aclweb.org/anthology/D17-1042} {A causal framework for
  explaining the predictions of black-box sequence-to-sequence models}.
\newblock In \emph{Proceedings of the 2017 Conference on Empirical Methods in
  Natural Language Processing}, pages 412--421. Association for Computational
  Linguistics.

\bibitem[{Alzantot et~al.(2018)Alzantot, Sharma, Elgohary, Ho, Srivastava, and
  Chang}]{alzantot2018generating}
Moustafa Alzantot, Yash Sharma, Ahmed Elgohary, Bo-Jhang Ho, Mani Srivastava,
  and Kai-Wei Chang. 2018.
\newblock \href {http://aclweb.org/anthology/D18-1316} {{Generating Natural
  Language Adversarial Examples}}.
\newblock In \emph{Proceedings of the 2018 Conference on Empirical Methods in
  Natural Language Processing}, pages 2890--2896. Association for Computational
  Linguistics.

\bibitem[{Arras et~al.(2017{\natexlab{a}})Arras, Horn, Montavon, M{\"u}ller,
  and Samek}]{10.1371/journal.pone.0181142}
Leila Arras, Franziska Horn, Gr{\'e}goire Montavon, Klaus-Robert M{\"u}ller,
  and Wojciech Samek. 2017{\natexlab{a}}.
\newblock \href {https://doi.org/10.1371/journal.pone.0181142} {{"What is
  relevant in a text document?": An interpretable machine learning approach}}.
\newblock \emph{PLOS ONE}, 12(8):1--23.

\bibitem[{Arras et~al.(2017{\natexlab{b}})Arras, Montavon, M{\"u}ller, and
  Samek}]{W17-5221}
Leila Arras, Gr{\'e}goire Montavon, Klaus-Robert M{\"u}ller, and Wojciech
  Samek. 2017{\natexlab{b}}.
\newblock \href {http://aclweb.org/anthology/W17-5221} {{Explaining Recurrent
  Neural Network Predictions in Sentiment Analysis}}.
\newblock In \emph{Proceedings of the 8th Workshop on Computational Approaches
  to Subjectivity, Sentiment and Social Media Analysis}, pages 159--168.
  Association for Computational Linguistics.

\bibitem[{Artetxe et~al.(2018)Artetxe, Labaka, Lopez-Gazpio, and
  Agirre}]{K18-1028}
Mikel Artetxe, Gorka Labaka, Inigo Lopez-Gazpio, and Eneko Agirre. 2018.
\newblock \href {http://aclweb.org/anthology/K18-1028} {{Uncovering Divergent
  Linguistic Information in Word Embeddings with Lessons for Intrinsic and
  Extrinsic Evaluation}}.
\newblock In \emph{Proceedings of the 22nd Conference on Computational Natural
  Language Learning}, pages 282--291. Association for Computational
  Linguistics.

\bibitem[{Aubakirova and Bansal(2016)}]{D16-1216}
Malika Aubakirova and Mohit Bansal. 2016.
\newblock \href {https://doi.org/10.18653/v1/D16-1216} {{Interpreting Neural
  Networks to Improve Politeness Comprehension}}.
\newblock In \emph{Proceedings of the 2016 Conference on Empirical Methods in
  Natural Language Processing}, pages 2035--2041. Association for Computational
  Linguistics.

\bibitem[{Bahdanau et~al.(2014)Bahdanau, Cho, and Bengio}]{bahdanau2014neural}
Dzmitry Bahdanau, Kyunghyun Cho, and Yoshua Bengio. 2014.
\newblock {Neural Machine Translation by Jointly Learning to Align and
  Translate}.
\newblock \emph{arXiv preprint arXiv:1409.0473v7}.

\bibitem[{Bau et~al.(2018)Bau, Belinkov, Sajjad, Durrani, Dalvi, and
  Glass}]{bau2018identifying}
Anthony Bau, Yonatan Belinkov, Hassan Sajjad, Nadir Durrani, Fahim Dalvi, and
  James Glass. 2018.
\newblock {Identifying and Controlling Important Neurons in Neural Machine
  Translation}.
\newblock \emph{arXiv preprint arXiv:1811.01157v1}.

\bibitem[{Bawden et~al.(2018)Bawden, Sennrich, Birch, and Haddow}]{N18-1118}
Rachel Bawden, Rico Sennrich, Alexandra Birch, and Barry Haddow. 2018.
\newblock \href {http://aclweb.org/anthology/N18-1118} {{Evaluating Discourse
  Phenomena in Neural Machine Translation}}.
\newblock In \emph{Proceedings of the 2018 Conference of the North American
  Chapter of the Association for Computational Linguistics: Human Language
  Technologies, Volume 1 (Long Papers)}, pages 1304--1313. Association for
  Computational Linguistics.

\bibitem[{Belinkov(2018)}]{belinkov:2018:phdthesis}
Yonatan Belinkov. 2018.
\newblock \emph{On Internal Language Representations in Deep Learning: An
  Analysis of Machine Translation and Speech Recognition}.
\newblock Ph.D. thesis, Massachusetts Institute of Technology.

\bibitem[{Belinkov and Bisk(2018)}]{belinkov:2018:ICLR}
Yonatan Belinkov and Yonatan Bisk. 2018.
\newblock {Synthetic and Natural Noise Both Break Neural Machine Translation}.
\newblock In \emph{International Conference on Learning Representations
  (ICLR)}.

\bibitem[{Belinkov et~al.(2017{\natexlab{a}})Belinkov, Durrani, Dalvi, Sajjad,
  and Glass}]{belinkov:2017:acl}
Yonatan Belinkov, Nadir Durrani, Fahim Dalvi, Hassan Sajjad, and James Glass.
  2017{\natexlab{a}}.
\newblock \href {https://doi.org/10.18653/v1/P17-1080} {{What do Neural Machine
  Translation Models Learn about Morphology?}}
\newblock In \emph{Proceedings of the 55th Annual Meeting of the Association
  for Computational Linguistics (Volume 1: Long Papers)}, pages 861--872.
  Association for Computational Linguistics.

\bibitem[{Belinkov and Glass(2017)}]{belinkov:2017:nips}
Yonatan Belinkov and James Glass. 2017.
\newblock \href
  {http://papers.nips.cc/paper/6838-analyzing-hidden-representations-in-end-to-end-automatic-speech-recognition-systems.pdf}
  {{Analyzing Hidden Representations in End-to-End Automatic Speech Recognition
  Systems}}.
\newblock In I.~Guyon, U.~V. Luxburg, S.~Bengio, H.~Wallach, R.~Fergus,
  S.~Vishwanathan, and R.~Garnett, editors, \emph{Advances in Neural
  Information Processing Systems 30}, pages 2441--2451. Curran Associates, Inc.

\bibitem[{Belinkov et~al.(2017{\natexlab{b}})Belinkov, M{\`a}rquez, Sajjad,
  Durrani, Dalvi, and Glass}]{belinkov:2017:ijcnlp}
Yonatan Belinkov, Llu{\'i}s M{\`a}rquez, Hassan Sajjad, Nadir Durrani, Fahim
  Dalvi, and James Glass. 2017{\natexlab{b}}.
\newblock \href {http://aclweb.org/anthology/I17-1001} {{Evaluating Layers of
  Representation in Neural Machine Translation on Part-of-Speech and Semantic
  Tagging Tasks}}.
\newblock In \emph{Proceedings of the Eighth International Joint Conference on
  Natural Language Processing (Volume 1: Long Papers)}, pages 1--10. Asian
  Federation of Natural Language Processing.

\bibitem[{Bernardy(2018)}]{bernardy2018can}
Jean-Philippe Bernardy. 2018.
\newblock {Can Recurrent Neural Networks Learn Nested Recursion?}
\newblock \emph{LiLT (Linguistic Issues in Language Technology)}, 16(1).

\bibitem[{Bisazza and Tump(2018)}]{D18-1313}
Arianna Bisazza and Clara Tump. 2018.
\newblock \href {http://aclweb.org/anthology/D18-1313} {{The Lazy Encoder: A
  Fine-Grained Analysis of the Role of Morphology in Neural Machine
  Translation}}.
\newblock In \emph{Proceedings of the 2018 Conference on Empirical Methods in
  Natural Language Processing}, pages 2871--2876. Association for Computational
  Linguistics.

\bibitem[{Blevins et~al.(2018)Blevins, Levy, and Zettlemoyer}]{P18-2003}
Terra Blevins, Omer Levy, and Luke Zettlemoyer. 2018.
\newblock \href {http://aclweb.org/anthology/P18-2003} {{Deep RNNs Encode Soft
  Hierarchical Syntax}}.
\newblock In \emph{Proceedings of the 56th Annual Meeting of the Association
  for Computational Linguistics (Volume 2: Short Papers)}, pages 14--19.
  Association for Computational Linguistics.

\bibitem[{Bod{\'e}n and Wiles(2002)}]{boden2002learning}
Mikael Bod{\'e}n and Janet Wiles. 2002.
\newblock On learning context-free and context-sensitive languages.
\newblock \emph{IEEE Transactions on Neural Networks}, 13(2):491--493.

\bibitem[{Bowman et~al.(2015)Bowman, Angeli, Potts, and Manning}]{D15-1075}
Samuel~R. Bowman, Gabor Angeli, Christopher Potts, and Christopher~D. Manning.
  2015.
\newblock \href {https://doi.org/10.18653/v1/D15-1075} {A large annotated
  corpus for learning natural language inference}.
\newblock In \emph{Proceedings of the 2015 Conference on Empirical Methods in
  Natural Language Processing}, pages 632--642. Association for Computational
  Linguistics.

\bibitem[{Bruni et~al.(2012)Bruni, Boleda, Baroni, and Tran}]{P12-1015}
Elia Bruni, Gemma Boleda, Marco Baroni, and Nam~Khanh Tran. 2012.
\newblock \href {http://aclweb.org/anthology/P12-1015} {{Distributional
  Semantics in Technicolor}}.
\newblock In \emph{Proceedings of the 50th Annual Meeting of the Association
  for Computational Linguistics (Volume 1: Long Papers)}, pages 136--145.
  Association for Computational Linguistics.

\bibitem[{Brunner et~al.(2017)Brunner, Wang, Wattenhofer, and
  Weigelt}]{brunner2018natural}
Gino Brunner, Yuyi Wang, Roger Wattenhofer, and Michael Weigelt. 2017.
\newblock {Natural Language Multitasking: Analyzing and Improving Syntactic
  Saliency of Hidden Representations}.
\newblock \emph{The 31st Annual Conference on Neural Information Processing
  (NIPS) - Workshop on Learning Disentangled Features: from Perception to
  Control}.

\bibitem[{Burchardt et~al.(2017)Burchardt, Macketanz, Dehdari, Heigold, Peter,
  and Williams}]{burchardt2017linguistic}
Aljoscha Burchardt, Vivien Macketanz, Jon Dehdari, Georg Heigold, Jan-Thorsten
  Peter, and Philip Williams. 2017.
\newblock {A Linguistic Evaluation of Rule-Based, Phrase-Based, and Neural MT
  Engines}.
\newblock \emph{The Prague Bulletin of Mathematical Linguistics},
  108(1):159--170.

\bibitem[{Burlot and Yvon(2017)}]{W17-4705}
Franck Burlot and Fran{\c{c}}ois Yvon. 2017.
\newblock \href {http://aclweb.org/anthology/W17-4705} {{Evaluating the
  morphological competence of Machine Translation Systems}}.
\newblock In \emph{Proceedings of the Second Conference on Machine
  Translation}, pages 43--55. Association for Computational Linguistics.

\bibitem[{Casey(1996)}]{casey1996dynamics}
Mike Casey. 1996.
\newblock \href {https://doi.org/10.1162/neco.1996.8.6.1135} {{The Dynamics of
  Discrete-Time Computation, with Application to Recurrent Neural Networks and
  Finite State Machine Extraction}}.
\newblock \emph{Neural computation}, 8(6):1135--1178.

\bibitem[{Cer et~al.(2017)Cer, Diab, Agirre, Lopez-Gazpio, and
  Specia}]{S17-2001}
Daniel Cer, Mona Diab, Eneko Agirre, Inigo Lopez-Gazpio, and Lucia Specia.
  2017.
\newblock \href {https://doi.org/10.18653/v1/S17-2001} {{SemEval-2017 Task 1:
  Semantic Textual Similarity Multilingual and Crosslingual Focused
  Evaluation}}.
\newblock In \emph{Proceedings of the 11th International Workshop on Semantic
  Evaluation (SemEval-2017)}, pages 1--14. Association for Computational
  Linguistics.

\bibitem[{Chaabouni et~al.(2017)Chaabouni, Dunbar, Zeghidour, and
  Dupoux}]{chaabouni2017learning}
Rahma Chaabouni, Ewan Dunbar, Neil Zeghidour, and Emmanuel Dupoux. 2017.
\newblock Learning weakly supervised multimodal phoneme embeddings.
\newblock In \emph{Interspeech 2017}.

\bibitem[{Chalup and Blair(2003)}]{chalup2003incremental}
Stephan~K. Chalup and Alan~D. Blair. 2003.
\newblock \href {https://doi.org/10.1016/S0893-6080(03)00054-6} {{Incremental
  Training of First Order Recurrent Neural Networks to Predict a
  Context-sensitive Language}}.
\newblock \emph{Neural Networks}, 16(7):955--972.

\bibitem[{Chang et~al.(2009)Chang, Gerrish, Wang, Boyd-graber, and
  Blei}]{NIPS2009_3700}
Jonathan Chang, Sean Gerrish, Chong Wang, Jordan~L. Boyd-graber, and David~M.
  Blei. 2009.
\newblock \href
  {http://papers.nips.cc/paper/3700-reading-tea-leaves-how-humans-interpret-topic-models.pdf}
  {{Reading Tea Leaves: How Humans Interpret Topic Models}}.
\newblock In Y.~Bengio, D.~Schuurmans, J.~D. Lafferty, C.~K.~I. Williams, and
  A.~Culotta, editors, \emph{Advances in Neural Information Processing Systems
  22}, pages 288--296. Curran Associates, Inc.

\bibitem[{Chen et~al.(2018{\natexlab{a}})Chen, Zhang, Chen, Yi, and
  Hsieh}]{P18-1241}
Hongge Chen, Huan Zhang, Pin-Yu Chen, Jinfeng Yi, and Cho-Jui Hsieh.
  2018{\natexlab{a}}.
\newblock \href {http://aclweb.org/anthology/P18-1241} {Attacking visual
  language grounding with adversarial examples: A case study on neural image
  captioning}.
\newblock In \emph{Proceedings of the 56th Annual Meeting of the Association
  for Computational Linguistics (Volume 1: Long Papers)}, pages 2587--2597.
  Association for Computational Linguistics.

\bibitem[{Chen et~al.(2015)Chen, Qiu, Zhu, Wu, and Huang}]{D15-1092}
Xinchi Chen, Xipeng Qiu, Chenxi Zhu, Shiyu Wu, and Xuanjing Huang. 2015.
\newblock \href {https://doi.org/10.18653/v1/D15-1092} {{Sentence Modeling with
  Gated Recursive Neural Network}}.
\newblock In \emph{Proceedings of the 2015 Conference on Empirical Methods in
  Natural Language Processing}, pages 793--798. Association for Computational
  Linguistics.

\bibitem[{Chen et~al.(2018{\natexlab{b}})Chen, Gilroy, Maletti, May, and
  Knight}]{N18-1205}
Yining Chen, Sorcha Gilroy, Andreas Maletti, Jonathan May, and Kevin Knight.
  2018{\natexlab{b}}.
\newblock \href {http://aclweb.org/anthology/N18-1205} {{Recurrent Neural
  Networks as Weighted Language Recognizers}}.
\newblock In \emph{Proceedings of the 2018 Conference of the North American
  Chapter of the Association for Computational Linguistics: Human Language
  Technologies, Volume 1 (Long Papers)}, pages 2261--2271. Association for
  Computational Linguistics.

\bibitem[{Cheng et~al.(2018)Cheng, Yi, Zhang, Chen, and
  Hsieh}]{cheng2018seq2sick}
Minhao Cheng, Jinfeng Yi, Huan Zhang, Pin-Yu Chen, and Cho-Jui Hsieh. 2018.
\newblock {Seq2Sick: Evaluating the Robustness of Sequence-to-Sequence Models
  with Adversarial Examples}.
\newblock \emph{arXiv preprint arXiv:1803.01128v1}.

\bibitem[{Chrupa{\l}a et~al.(2017)Chrupa{\l}a, Gelderloos, and
  Alishahi}]{chrupala2017representations}
Grzegorz Chrupa{\l}a, Lieke Gelderloos, and Afra Alishahi. 2017.
\newblock \href {https://doi.org/10.18653/v1/P17-1057} {{Representations of
  language in a model of visually grounded speech signal}}.
\newblock In \emph{Proceedings of the 55th Annual Meeting of the Association
  for Computational Linguistics (Volume 1: Long Papers)}, pages 613--622.
  Association for Computational Linguistics.

\bibitem[{C{\'i}fka and Bojar(2018)}]{P18-1126}
Ond{\v{r}}ej C{\'i}fka and Ond{\v{r}}ej Bojar. 2018.
\newblock \href {http://aclweb.org/anthology/P18-1126} {{Are BLEU and Meaning
  Representation in Opposition?}}
\newblock In \emph{Proceedings of the 56th Annual Meeting of the Association
  for Computational Linguistics (Volume 1: Long Papers)}, pages 1362--1371.
  Association for Computational Linguistics.

\bibitem[{Conneau et~al.(2018)Conneau, Kruszewski, Lample, Barrault, and
  Baroni}]{conneau2018you}
Alexis Conneau, Germ{\'a}n Kruszewski, Guillaume Lample, Lo{\"i}c Barrault, and
  Marco Baroni. 2018.
\newblock \href {http://aclweb.org/anthology/P18-1198} {What you can cram into
  a single {\$}{\&}!{\#}* vector: Probing sentence embeddings for linguistic
  properties}.
\newblock In \emph{Proceedings of the 56th Annual Meeting of the Association
  for Computational Linguistics (Volume 1: Long Papers)}, pages 2126--2136.
  Association for Computational Linguistics.

\bibitem[{Cooper et~al.(1996)Cooper, Crouch, van Eijck, Fox, van Genabith,
  Jaspars, Kamp, Milward, Pinkal, Poesio, Pulman, Briscoe, Maier, and
  Konrad}]{cooper1996using}
Robin Cooper, Dick Crouch, Jan van Eijck, Chris Fox, Josef van Genabith, Jan
  Jaspars, Hans Kamp, David Milward, Manfred Pinkal, Massimo Poesio, Steve
  Pulman, Ted Briscoe, Holger Maier, and Karsten Konrad. 1996.
\newblock Using the framework.
\newblock Technical report, The FraCaS Consortium.

\bibitem[{Dalvi et~al.(2019{\natexlab{a}})Dalvi, Durrani, Sajjad, Belinkov,
  Bau, and Glass}]{dalvi:2019:AAAI}
Fahim Dalvi, Nadir Durrani, Hassan Sajjad, Yonatan Belinkov, D.~Anthony Bau,
  and James Glass. 2019{\natexlab{a}}.
\newblock {What Is One Grain of Sand in the Desert? Analyzing Individual
  Neurons in Deep NLP Models}.
\newblock In \emph{Proceedings of the Thirty-Third AAAI Conference on
  Artificial Intelligence (AAAI)}.

\bibitem[{Dalvi et~al.(2017)Dalvi, Durrani, Sajjad, Belinkov, and
  Vogel}]{dalvi:2017:ijcnlp}
Fahim Dalvi, Nadir Durrani, Hassan Sajjad, Yonatan Belinkov, and Stephan Vogel.
  2017.
\newblock \href {http://aclweb.org/anthology/I17-1015} {{Understanding and
  Improving Morphological Learning in the Neural Machine Translation Decoder}}.
\newblock In \emph{Proceedings of the Eighth International Joint Conference on
  Natural Language Processing (Volume 1: Long Papers)}, pages 142--151. Asian
  Federation of Natural Language Processing.

\bibitem[{Dalvi et~al.(2019{\natexlab{b}})Dalvi, Nortonsmith, Bau, Belinkov,
  Sajjad, Durrani, and Glass}]{dalvi:2019:AAAI:demo}
Fahim Dalvi, Avery Nortonsmith, D.~Anthony Bau, Yonatan Belinkov, Hassan
  Sajjad, Nadir Durrani, and James Glass. 2019{\natexlab{b}}.
\newblock {NeuroX: A Toolkit for Analyzing Individual Neurons in Neural
  Networks}.
\newblock In \emph{Proceedings of the Thirty-Third AAAI Conference on
  Artificial Intelligence (AAAI): Demonstrations Track}.

\bibitem[{Das et~al.(1992)Das, Giles, and Sun}]{das1992learning}
Sreerupa Das, C.~Lee Giles, and Guo-Zheng Sun. 1992.
\newblock {Learning Context-free Grammars: Capabilities and Limitations of a
  Recurrent Neural Network with an External Stack Memory}.
\newblock In \emph{Proceedings of The Fourteenth Annual Conference of Cognitive
  Science Society. Indiana University}, page~14.

\bibitem[{Dasgupta et~al.(2018)Dasgupta, Guo, Stuhlm{\"u}ller, Gershman, and
  Goodman}]{dasgupta2018evaluating}
Ishita Dasgupta, Demi Guo, Andreas Stuhlm{\"u}ller, Samuel~J. Gershman, and
  Noah~D. Goodman. 2018.
\newblock {Evaluating Compositionality in Sentence Embeddings}.
\newblock \emph{arXiv preprint arXiv:1802.04302v2}.

\bibitem[{Dharmaretnam and Fyshe(2018)}]{N18-2122}
Dhanush Dharmaretnam and Alona Fyshe. 2018.
\newblock \href {http://aclweb.org/anthology/N18-2122} {{The Emergence of
  Semantics in Neural Network Representations of Visual Information}}.
\newblock In \emph{Proceedings of the 2018 Conference of the North American
  Chapter of the Association for Computational Linguistics: Human Language
  Technologies, Volume 2 (Short Papers)}, pages 776--780. Association for
  Computational Linguistics.

\bibitem[{Ding et~al.(2017)Ding, Liu, Luan, and Sun}]{P17-1106}
Yanzhuo Ding, Yang Liu, Huanbo Luan, and Maosong Sun. 2017.
\newblock \href {https://doi.org/10.18653/v1/P17-1106} {{Visualizing and
  Understanding Neural Machine Translation}}.
\newblock In \emph{Proceedings of the 55th Annual Meeting of the Association
  for Computational Linguistics (Volume 1: Long Papers)}, pages 1150--1159.
  Association for Computational Linguistics.

\bibitem[{Doshi-Velez and Kim(2017)}]{DoshiKim2017Interpretability}
Finale Doshi-Velez and Been Kim. 2017.
\newblock {Towards A Rigorous Science of Interpretable Machine Learning}.
\newblock In \emph{arXiv preprint arXiv:1702.08608v2}.

\bibitem[{Doshi-Velez et~al.(2017)Doshi-Velez, Kortz, Budish, Bavitz, Gershman,
  O'Brien, Shieber, Waldo, Weinberger, and Wood}]{doshi2017accountability}
Finale Doshi-Velez, Mason Kortz, Ryan Budish, Chris Bavitz, Sam Gershman, David
  O'Brien, Stuart Shieber, James Waldo, David Weinberger, and Alexandra Wood.
  2017.
\newblock {Accountability of AI Under the Law: The Role of Explanation}.
\newblock \emph{Berkman Center Publication Forthcoming}.

\bibitem[{Drexler and Glass(2017)}]{Drexler2017AnalysisOA}
Jennifer Drexler and James Glass. 2017.
\newblock {Analysis of Audio-Visual Features for Unsupervised Speech
  Recognition}.
\newblock In \emph{International Workshop on Grounding Language Understanding}.

\bibitem[{Ebrahimi et~al.(2018{\natexlab{a}})Ebrahimi, Lowd, and
  Dou}]{C18-1055}
Javid Ebrahimi, Daniel Lowd, and Dejing Dou. 2018{\natexlab{a}}.
\newblock \href {http://aclweb.org/anthology/C18-1055} {{On Adversarial
  Examples for Character-Level Neural Machine Translation}}.
\newblock In \emph{Proceedings of the 27th International Conference on
  Computational Linguistics}, pages 653--663. Association for Computational
  Linguistics.

\bibitem[{Ebrahimi et~al.(2018{\natexlab{b}})Ebrahimi, Rao, Lowd, and
  Dou}]{P18-2006}
Javid Ebrahimi, Anyi Rao, Daniel Lowd, and Dejing Dou. 2018{\natexlab{b}}.
\newblock \href {http://aclweb.org/anthology/P18-2006} {{HotFlip: White-Box
  Adversarial Examples for Text Classification}}.
\newblock In \emph{Proceedings of the 56th Annual Meeting of the Association
  for Computational Linguistics (Volume 2: Short Papers)}, pages 31--36.
  Association for Computational Linguistics.

\bibitem[{Elkahky et~al.(2018)Elkahky, Webster, Andor, and Pitler}]{D18-1277}
Ali Elkahky, Kellie Webster, Daniel Andor, and Emily Pitler. 2018.
\newblock \href {http://aclweb.org/anthology/D18-1277} {{A Challenge Set and
  Methods for Noun-Verb Ambiguity}}.
\newblock In \emph{Proceedings of the 2018 Conference on Empirical Methods in
  Natural Language Processing}, pages 2562--2572. Association for Computational
  Linguistics.

\bibitem[{Elloumi et~al.(2018)Elloumi, Besacier, Galibert, and
  Lecouteux}]{elloumi2018analyzing}
Zied Elloumi, Laurent Besacier, Olivier Galibert, and Benjamin Lecouteux. 2018.
\newblock \href {http://aclweb.org/anthology/W18-5402} {{Analyzing Learned
  Representations of a Deep ASR Performance Prediction Model}}.
\newblock In \emph{Proceedings of the 2018 EMNLP Workshop BlackboxNLP:
  Analyzing and Interpreting Neural Networks for NLP}, pages 9--15. Association
  for Computational Linguistics.

\bibitem[{Elman(1989)}]{elman1989representation}
Jeffrey~L. Elman. 1989.
\newblock {Representation and Structure in Connectionist Models}.
\newblock Technical report, University of California, San Diego, Center for
  Research in Language.

\bibitem[{Elman(1990)}]{elman1990finding}
Jeffrey~L. Elman. 1990.
\newblock {Finding Structure in Time}.
\newblock \emph{Cognitive science}, 14(2):179--211.

\bibitem[{Elman(1991)}]{elman1991distributed}
Jeffrey~L. Elman. 1991.
\newblock Distributed representations, simple recurrent networks, and
  grammatical structure.
\newblock \emph{Machine learning}, 7(2-3):195--225.

\bibitem[{Ettinger et~al.(2016)Ettinger, Elgohary, and Resnik}]{W16-2524}
Allyson Ettinger, Ahmed Elgohary, and Philip Resnik. 2016.
\newblock \href {https://doi.org/10.18653/v1/W16-2524} {Probing for semantic
  evidence of composition by means of simple classification tasks}.
\newblock In \emph{Proceedings of the 1st Workshop on Evaluating Vector-Space
  Representations for NLP}, pages 134--139. Association for Computational
  Linguistics.

\bibitem[{Faruqui et~al.(2016)Faruqui, Tsvetkov, Rastogi, and
  Dyer}]{Faruqui:repeval:16}
Manaal Faruqui, Yulia Tsvetkov, Pushpendre Rastogi, and Chris Dyer. 2016.
\newblock \href {http://arxiv.org/pdf/1605.02276v1.pdf} {{Problems With
  Evaluation of Word Embeddings Using Word Similarity Tasks}}.
\newblock In \emph{Proc. of the 1st Workshop on Evaluating Vector Space
  Representations for NLP}.

\bibitem[{Faruqui et~al.(2015)Faruqui, Tsvetkov, Yogatama, Dyer, and
  Smith}]{P15-1144}
Manaal Faruqui, Yulia Tsvetkov, Dani Yogatama, Chris Dyer, and Noah~A. Smith.
  2015.
\newblock \href {https://doi.org/10.3115/v1/P15-1144} {{Sparse Overcomplete
  Word Vector Representations}}.
\newblock In \emph{Proceedings of the 53rd Annual Meeting of the Association
  for Computational Linguistics and the 7th International Joint Conference on
  Natural Language Processing (Volume 1: Long Papers)}, pages 1491--1500.
  Association for Computational Linguistics.

\bibitem[{Feng et~al.(2018)Feng, Wallace, Grissom~II, Iyyer, Rodriguez, and
  Boyd-Graber}]{FengRAWR2018}
Shi Feng, Eric Wallace, Alvin Grissom~II, Mohit Iyyer, Pedro Rodriguez, and
  Jordan Boyd-Graber. 2018.
\newblock \href {http://aclweb.org/anthology/D18-1407} {{Pathologies of Neural
  Models Make Interpretations Difficult}}.
\newblock In \emph{Proceedings of the 2018 Conference on Empirical Methods in
  Natural Language Processing}, pages 3719--3728. Association for Computational
  Linguistics.

\bibitem[{Finkelstein et~al.(2002)Finkelstein, Gabrilovich, Matias, Rivlin,
  Solan, Wolfman, and Ruppin}]{finkelstein2002placing}
Lev Finkelstein, Evgeniy Gabrilovich, Yossi Matias, Ehud Rivlin, Zach Solan,
  Gadi Wolfman, and Eytan Ruppin. 2002.
\newblock \href {https://doi.org/10.1145/503104.503110} {{Placing Search in
  Context: The Concept Revisited}}.
\newblock \emph{ACM Transactions on information systems}, 20(1):116--131.

\bibitem[{Frank et~al.(2013)Frank, Mathis, and
  Badecker}]{doi:10.1080/10489223.2013.796950}
Robert Frank, Donald Mathis, and William Badecker. 2013.
\newblock \href {https://doi.org/10.1080/10489223.2013.796950} {{The
  Acquisition of Anaphora by Simple Recurrent Networks}}.
\newblock \emph{Language Acquisition}, 20(3):181--227.

\bibitem[{Freeman et~al.(2018)Freeman, Merriman, Aggarwal, Beaver, and
  Mueen}]{freeman2018paying}
Cynthia Freeman, Jonathan Merriman, Abhinav Aggarwal, Ian Beaver, and Abdullah
  Mueen. 2018.
\newblock {Paying Attention to Attention: Highlighting Influential Samples in
  Sequential Analysis}.
\newblock \emph{arXiv preprint arXiv:1808.02113v1}.

\bibitem[{Fyshe et~al.(2015)Fyshe, Wehbe, Talukdar, Murphy, and
  Mitchell}]{N15-1004}
Alona Fyshe, Leila Wehbe, Partha~P. Talukdar, Brian Murphy, and Tom~M.
  Mitchell. 2015.
\newblock \href {https://doi.org/10.3115/v1/N15-1004} {{A Compositional and
  Interpretable Semantic Space}}.
\newblock In \emph{Proceedings of the 2015 Conference of the North American
  Chapter of the Association for Computational Linguistics: Human Language
  Technologies}, pages 32--41. Association for Computational Linguistics.

\bibitem[{Gaddy et~al.(2018)Gaddy, Stern, and Klein}]{N18-1091}
David Gaddy, Mitchell Stern, and Dan Klein. 2018.
\newblock \href {http://aclweb.org/anthology/N18-1091} {{What's Going On in
  Neural Constituency Parsers? An Analysis}}.
\newblock In \emph{Proceedings of the 2018 Conference of the North American
  Chapter of the Association for Computational Linguistics: Human Language
  Technologies, Volume 1 (Long Papers)}, pages 999--1010. Association for
  Computational Linguistics.

\bibitem[{Ganesh et~al.(2017)Ganesh, Gupta, and
  Varma}]{Ganesh:2017:IST:3110025.3110083}
J.~Ganesh, Manish Gupta, and Vasudeva Varma. 2017.
\newblock \href {https://doi.org/10.1145/3110025.3110083} {{Interpretation of
  Semantic Tweet Representations}}.
\newblock In \emph{Proceedings of the 2017 IEEE/ACM International Conference on
  Advances in Social Networks Analysis and Mining 2017}, ASONAM '17, pages
  95--102, New York, NY, USA. ACM.

\bibitem[{Gao et~al.(2018)Gao, Lanchantin, Soffa, and Qi}]{gao2018black}
Ji~Gao, Jack Lanchantin, Mary~Lou Soffa, and Yanjun Qi. 2018.
\newblock {Black-box Generation of Adversarial Text Sequences to Evade Deep
  Learning Classifiers}.
\newblock \emph{arXiv preprint arXiv:1801.04354v5}.

\bibitem[{Gelderloos and Chrupa{\l}a(2016)}]{gelderloos-chrupala:2016:COLING}
Lieke Gelderloos and Grzegorz Chrupa{\l}a. 2016.
\newblock \href {http://aclweb.org/anthology/C16-1124} {{From phonemes to
  images: Levels of representation in a recurrent neural model of
  visually-grounded language learning}}.
\newblock In \emph{Proceedings of COLING 2016, the 26th International
  Conference on Computational Linguistics: Technical Papers}, pages 1309--1319,
  Osaka, Japan. The COLING 2016 Organizing Committee.

\bibitem[{Gers and Schmidhuber(2001)}]{gers2001lstm}
Felix~A. Gers and J{\"u}rgen Schmidhuber. 2001.
\newblock \href {https://doi.org/10.1109/72.963769} {{LSTM Recurrent Networks
  Learn Simple Context-Free and Context-Sensitive Languages}}.
\newblock \emph{IEEE Transactions on Neural Networks}, 12(6):1333--1340.

\bibitem[{Gerz et~al.(2016)Gerz, Vuli{\'{c}}, Hill, Reichart, and
  Korhonen}]{D16-1235}
Daniela Gerz, Ivan Vuli{\'{c}}, Felix Hill, Roi Reichart, and Anna Korhonen.
  2016.
\newblock \href {https://doi.org/10.18653/v1/D16-1235} {{SimVerb-3500: A
  Large-Scale Evaluation Set of Verb Similarity}}.
\newblock In \emph{Proceedings of the 2016 Conference on Empirical Methods in
  Natural Language Processing}, pages 2173--2182. Association for Computational
  Linguistics.

\bibitem[{Ghader and Monz(2017)}]{I17-1004}
Hamidreza Ghader and Christof Monz. 2017.
\newblock \href {http://aclweb.org/anthology/I17-1004} {{What does Attention in
  Neural Machine Translation Pay Attention to?}}
\newblock In \emph{Proceedings of the Eighth International Joint Conference on
  Natural Language Processing (Volume 1: Long Papers)}, pages 30--39. Asian
  Federation of Natural Language Processing.

\bibitem[{Ghaeini et~al.(2018)Ghaeini, Fern, and
  Tadepalli}]{ghaeini2018interpreting}
Reza Ghaeini, Xiaoli Fern, and Prasad Tadepalli. 2018.
\newblock \href {http://aclweb.org/anthology/D18-1537} {{Interpreting Recurrent
  and Attention-Based Neural Models: A Case Study on Natural Language
  Inference}}.
\newblock In \emph{Proceedings of the 2018 Conference on Empirical Methods in
  Natural Language Processing}, pages 4952--4957. Association for Computational
  Linguistics.

\bibitem[{Giulianelli et~al.(2018)Giulianelli, Harding, Mohnert, Hupkes, and
  Zuidema}]{giulianelli2018under}
Mario Giulianelli, Jack Harding, Florian Mohnert, Dieuwke Hupkes, and Willem
  Zuidema. 2018.
\newblock \href {http://aclweb.org/anthology/W18-5426} {{Under the Hood: Using
  Diagnostic Classifiers to Investigate and Improve how Language Models Track
  Agreement Information}}.
\newblock In \emph{Proceedings of the 2018 EMNLP Workshop BlackboxNLP:
  Analyzing and Interpreting Neural Networks for NLP}, pages 240--248.
  Association for Computational Linguistics.

\bibitem[{Glockner et~al.(2018)Glockner, Shwartz, and Goldberg}]{P18-2103}
Max Glockner, Vered Shwartz, and Yoav Goldberg. 2018.
\newblock \href {http://aclweb.org/anthology/P18-2103} {{Breaking NLI Systems
  with Sentences that Require Simple Lexical Inferences}}.
\newblock In \emph{Proceedings of the 56th Annual Meeting of the Association
  for Computational Linguistics (Volume 2: Short Papers)}, pages 650--655.
  Association for Computational Linguistics.

\bibitem[{Godin et~al.(2018)Godin, Demuynck, Dambre, De~Neve, and
  Demeester}]{godin2018explaining}
Fr{\'e}deric Godin, Kris Demuynck, Joni Dambre, Wesley De~Neve, and Thomas
  Demeester. 2018.
\newblock \href {http://aclweb.org/anthology/D18-1365} {{Explaining
  Character-Aware Neural Networks for Word-Level Prediction: Do They Discover
  Linguistic Rules?}}
\newblock In \emph{Proceedings of the 2018 Conference on Empirical Methods in
  Natural Language Processing}, pages 3275--3284. Association for Computational
  Linguistics.

\bibitem[{Goldberg(2017)}]{goldberg2017neural}
Yoav Goldberg. 2017.
\newblock \emph{{Neural Network methods for Natural Language Processing}},
  volume~10 of \emph{Synthesis Lectures on Human Language Technologies}.
\newblock Morgan \& Claypool Publishers.

\bibitem[{Goodfellow et~al.(2016)Goodfellow, Bengio, and
  Courville}]{Goodfellow-et-al-2016}
Ian Goodfellow, Yoshua Bengio, and Aaron Courville. 2016.
\newblock \emph{Deep Learning}.
\newblock MIT Press.
\newblock \url{http://www.deeplearningbook.org}.

\bibitem[{Goodfellow et~al.(2014)Goodfellow, Pouget-Abadie, Mirza, Xu,
  Warde-Farley, Ozair, Courville, and Bengio}]{goodfellow2014generative}
Ian Goodfellow, Jean Pouget-Abadie, Mehdi Mirza, Bing Xu, David Warde-Farley,
  Sherjil Ozair, Aaron Courville, and Yoshua Bengio. 2014.
\newblock {Generative Adversarial Nets}.
\newblock In \emph{Advances in neural information processing systems}, pages
  2672--2680.

\bibitem[{Goodfellow et~al.(2015)Goodfellow, Shlens, and
  Szegedy}]{goodfellow2014explaining}
Ian~J. Goodfellow, Jonathon Shlens, and Christian Szegedy. 2015.
\newblock {Explaining and Harnessing Adversarial Examples}.
\newblock In \emph{International Conference on Learning Representations
  (ICLR)}.

\bibitem[{Gulordava et~al.(2018)Gulordava, Bojanowski, Grave, Linzen, and
  Baroni}]{gulordava2018colorless}
Kristina Gulordava, Piotr Bojanowski, Edouard Grave, Tal Linzen, and Marco
  Baroni. 2018.
\newblock \href {http://aclweb.org/anthology/N18-1108} {{Colorless Green
  Recurrent Networks Dream Hierarchically}}.
\newblock In \emph{Proceedings of the 2018 Conference of the North American
  Chapter of the Association for Computational Linguistics: Human Language
  Technologies, Volume 1 (Long Papers)}, pages 1195--1205. Association for
  Computational Linguistics.

\bibitem[{Gupta et~al.(2015)Gupta, Boleda, Baroni, and Pad{\'o}}]{D15-1002}
Abhijeet Gupta, Gemma Boleda, Marco Baroni, and Sebastian Pad{\'o}. 2015.
\newblock \href {https://doi.org/10.18653/v1/D15-1002} {Distributional vectors
  encode referential attributes}.
\newblock In \emph{Proceedings of the 2015 Conference on Empirical Methods in
  Natural Language Processing}, pages 12--21. Association for Computational
  Linguistics.

\bibitem[{Gupta and Sch{\"u}tze(2018)}]{gupta2018lisa}
Pankaj Gupta and Hinrich Sch{\"u}tze. 2018.
\newblock \href {http://aclweb.org/anthology/W18-5418} {{LISA: Explaining
  Recurrent Neural Network Judgments via Layer-wIse Semantic Accumulation and
  Example to Pattern Transformation}}.
\newblock In \emph{Proceedings of the 2018 EMNLP Workshop BlackboxNLP:
  Analyzing and Interpreting Neural Networks for NLP}, pages 154--164.
  Association for Computational Linguistics.

\bibitem[{Gururangan et~al.(2018)Gururangan, Swayamdipta, Levy, Schwartz,
  Bowman, and Smith}]{N18-2017}
Suchin Gururangan, Swabha Swayamdipta, Omer Levy, Roy Schwartz, Samuel Bowman,
  and Noah~A. Smith. 2018.
\newblock \href {http://aclweb.org/anthology/N18-2017} {{Annotation Artifacts
  in Natural Language Inference Data}}.
\newblock In \emph{Proceedings of the 2018 Conference of the North American
  Chapter of the Association for Computational Linguistics: Human Language
  Technologies, Volume 2 (Short Papers)}, pages 107--112. Association for
  Computational Linguistics.

\bibitem[{Harris(1990)}]{doi:10.1080/09540099008915660}
Catherine~L. Harris. 1990.
\newblock \href {https://doi.org/10.1080/09540099008915660} {{Connectionism and
  Cognitive Linguistics}}.
\newblock \emph{Connection Science}, 2(1-2):7--33.

\bibitem[{Harwath and Glass(2017)}]{harwath2017learning}
David Harwath and James Glass. 2017.
\newblock \href {https://doi.org/10.18653/v1/P17-1047} {{Learning Word-Like
  Units from Joint Audio-Visual Analysis}}.
\newblock In \emph{Proceedings of the 55th Annual Meeting of the Association
  for Computational Linguistics (Volume 1: Long Papers)}, pages 506--517.
  Association for Computational Linguistics.

\bibitem[{Heigold et~al.(2018)Heigold, Neumann, and van
  Genabith}]{heigold2017robust}
Georg Heigold, G{\"u}nter Neumann, and Josef van Genabith. 2018.
\newblock {How Robust Are Character-Based Word Embeddings in Tagging and MT
  Against Wrod Scramlbing or Randdm Nouse?}
\newblock In \emph{Proceedings of the 13th Conference of The Association for
  Machine Translation in the Americas (Volume 1: Research Track)}, pages
  68--79.

\bibitem[{Hill et~al.(2015)Hill, Reichart, and Korhonen}]{J15-4004}
Felix Hill, Roi Reichart, and Anna Korhonen. 2015.
\newblock \href {https://doi.org/10.1162/COLI_a_00237} {{SimLex-999: Evaluating
  Semantic Models With (Genuine) Similarity Estimation}}.
\newblock \emph{Computational Linguistics}, 41(4):665--695.

\bibitem[{Hupkes et~al.(2018)Hupkes, Veldhoen, and
  Zuidema}]{hupkes2017visualisation}
Dieuwke Hupkes, Sara Veldhoen, and Willem Zuidema. 2018.
\newblock Visualisation and 'diagnostic classifiers' reveal how recurrent and
  recursive neural networks process hierarchical structure.
\newblock \emph{Journal of Artificial Intelligence Research}, 61:907--926.

\bibitem[{Isabelle et~al.(2017)Isabelle, Cherry, and Foster}]{D17-1263}
Pierre Isabelle, Colin Cherry, and George Foster. 2017.
\newblock \href {http://aclweb.org/anthology/D17-1263} {{A Challenge Set
  Approach to Evaluating Machine Translation}}.
\newblock In \emph{Proceedings of the 2017 Conference on Empirical Methods in
  Natural Language Processing}, pages 2486--2496. Association for Computational
  Linguistics.

\bibitem[{Isabelle and Kuhn(2018)}]{isabelle2018challenge}
Pierre Isabelle and Roland Kuhn. 2018.
\newblock {A Challenge Set for French--> English Machine Translation}.
\newblock \emph{arXiv preprint arXiv:1806.02725v2}.

\bibitem[{Isahara(1995)}]{isahara1995jeida}
Hitoshi Isahara. 1995.
\newblock {JEIDA's test-sets for quality evaluation of MT systems-technical
  evaluation from the developer's point of view}.
\newblock In \emph{Proceedings of MT Summit V}.

\bibitem[{Iyyer et~al.(2018)Iyyer, Wieting, Gimpel, and Zettlemoyer}]{N18-1170}
Mohit Iyyer, John Wieting, Kevin Gimpel, and Luke Zettlemoyer. 2018.
\newblock \href {http://aclweb.org/anthology/N18-1170} {{Adversarial Example
  Generation with Syntactically Controlled Paraphrase Networks}}.
\newblock In \emph{Proceedings of the 2018 Conference of the North American
  Chapter of the Association for Computational Linguistics: Human Language
  Technologies, Volume 1 (Long Papers)}, pages 1875--1885. Association for
  Computational Linguistics.

\bibitem[{Jacovi et~al.(2018)Jacovi, Sar~Shalom, and Goldberg}]{W18-5408}
Alon Jacovi, Oren Sar~Shalom, and Yoav Goldberg. 2018.
\newblock \href {http://aclweb.org/anthology/W18-5408} {{Understanding
  Convolutional Neural Networks for Text Classification}}.
\newblock In \emph{Proceedings of the 2018 EMNLP Workshop BlackboxNLP:
  Analyzing and Interpreting Neural Networks for NLP}, pages 56--65.
  Association for Computational Linguistics.

\bibitem[{Jauregi~Unanue et~al.(2018)Jauregi~Unanue, Zare~Borzeshi, and
  Piccardi}]{unanue2018shared}
Inigo Jauregi~Unanue, Ehsan Zare~Borzeshi, and Massimo Piccardi. 2018.
\newblock \href {http://aclweb.org/anthology/W18-2702} {{A Shared Attention
  Mechanism for Interpretation of Neural Automatic Post-Editing Systems}}.
\newblock In \emph{Proceedings of the 2nd Workshop on Neural Machine
  Translation and Generation}, pages 11--17. Association for Computational
  Linguistics.

\bibitem[{Jia and Liang(2017)}]{jia-liang:2017:EMNLP2017}
Robin Jia and Percy Liang. 2017.
\newblock \href {http://aclweb.org/anthology/D17-1215} {Adversarial examples
  for evaluating reading comprehension systems}.
\newblock In \emph{Proceedings of the 2017 Conference on Empirical Methods in
  Natural Language Processing}, pages 2021--2031. Association for Computational
  Linguistics.

\bibitem[{Jozefowicz et~al.(2016)Jozefowicz, Vinyals, Schuster, Shazeer, and
  Wu}]{jozefowicz2016exploring}
Rafal Jozefowicz, Oriol Vinyals, Mike Schuster, Noam Shazeer, and Yonghui Wu.
  2016.
\newblock {Exploring the Limits of Language Modeling}.
\newblock \emph{arXiv preprint arXiv:1602.02410v2}.

\bibitem[{K{\'a}d{\'a}r et~al.(2017)K{\'a}d{\'a}r, Chrupa{\l}a, and
  Alishahi}]{kadar2016representation}
Akos K{\'a}d{\'a}r, Grzegorz Chrupa{\l}a, and Afra Alishahi. 2017.
\newblock \href {https://doi.org/10.1162/COLI_a_00300} {{Representation of
  Linguistic Form and Function in Recurrent Neural Networks}}.
\newblock \emph{Computational Linguistics}, 43(4):761--780.

\bibitem[{Karpathy et~al.(2015)Karpathy, Johnson, and
  Li}]{karpathy2015visualizing}
Andrej Karpathy, Justin Johnson, and Fei-Fei Li. 2015.
\newblock {Visualizing and Understanding Recurrent Networks}.
\newblock \emph{arXiv preprint arXiv:1506.02078v2}.

\bibitem[{Khandelwal et~al.(2018)Khandelwal, He, Qi, and Jurafsky}]{P18-1027}
Urvashi Khandelwal, He~He, Peng Qi, and Dan Jurafsky. 2018.
\newblock \href {http://aclweb.org/anthology/P18-1027} {{Sharp Nearby, Fuzzy
  Far Away: How Neural Language Models Use Context}}.
\newblock In \emph{Proceedings of the 56th Annual Meeting of the Association
  for Computational Linguistics (Volume 1: Long Papers)}, pages 284--294.
  Association for Computational Linguistics.

\bibitem[{King and Falkedal(1990)}]{C90-2037}
Margaret King and Kirsten Falkedal. 1990.
\newblock \href {http://www.aclweb.org/anthology/C90-2037} {{Using Test Suites
  in Evaluation of Machine Translation Systems}}.
\newblock In \emph{COLNG 1990 Volume 2: Papers presented to the 13th
  International Conference on Computational Linguistics}.

\bibitem[{Kiperwasser and Goldberg(2016)}]{Q16-1023}
Eliyahu Kiperwasser and Yoav Goldberg. 2016.
\newblock \href {http://aclweb.org/anthology/Q16-1023} {{Simple and Accurate
  Dependency Parsing Using Bidirectional LSTM Feature Representations}}.
\newblock \emph{Transactions of the Association for Computational Linguistics},
  4:313--327.

\bibitem[{Koh et~al.(2001)Koh, Maeng, Lee, Chae, and Choi}]{koh2001test}
Sungryong Koh, Jinee Maeng, Ji-Young Lee, Young-Sook Chae, and Key-Sun Choi.
  2001.
\newblock {A test suite for evaluation of English-to-Korean machine translation
  systems}.
\newblock In \emph{MT Summit Conference}.

\bibitem[{K\"{o}hn(2015)}]{kohn:2015:EMNLP}
Arne K\"{o}hn. 2015.
\newblock \href {http://aclweb.org/anthology/D15-1246} {{What's in an
  Embedding? Analyzing Word Embeddings through Multilingual Evaluation}}.
\newblock In \emph{Proceedings of the 2015 Conference on Empirical Methods in
  Natural Language Processing}, pages 2067--2073, Lisbon, Portugal. Association
  for Computational Linguistics.

\bibitem[{Kuleshov et~al.(2018)Kuleshov, Thakoor, Lau, and
  Ermon}]{kuleshov2018adversarial}
Volodymyr Kuleshov, Shantanu Thakoor, Tingfung Lau, and Stefano Ermon. 2018.
\newblock \href {https://openreview.net/forum?id=r1QZ3zbAZ} {{Adversarial
  Examples for Natural Language Classification Problems}}.

\bibitem[{Lake and Baroni(2018)}]{lake2018}
Brenden Lake and Marco Baroni. 2018.
\newblock \href {http://proceedings.mlr.press/v80/lake18a.html}
  {{Generalization without Systematicity: On the Compositional Skills of
  Sequence-to-Sequence Recurrent Networks}}.
\newblock In \emph{Proceedings of the 35th International Conference on Machine
  Learning}, volume~80 of \emph{Proceedings of Machine Learning Research},
  pages 2873--2882, Stockholmsm{\"a}ssan, Stockholm, Sweden. PMLR.

\bibitem[{Lehmann et~al.(1996)Lehmann, Oepen, Regnier-Prost, Netter, Lux,
  Klein, Falkedal, Fouvry, Estival, Dauphin, Compagnion, Baur, Balkan, and
  Arnold}]{C96-2120}
Sabine Lehmann, Stephan Oepen, Sylvie Regnier-Prost, Klaus Netter, Veronika
  Lux, Judith Klein, Kirsten Falkedal, Frederik Fouvry, Dominique Estival, Eva
  Dauphin, Herve Compagnion, Judith Baur, Lorna Balkan, and Doug Arnold. 1996.
\newblock \href {http://www.aclweb.org/anthology/C96-2120} {{TSNLP - Test
  Suites for Natural Language Processing}}.
\newblock In \emph{COLING 1996 Volume 2: The 16th International Conference on
  Computational Linguistics}.

\bibitem[{Lei et~al.(2016)Lei, Barzilay, and Jaakkola}]{D16-1011}
Tao Lei, Regina Barzilay, and Tommi Jaakkola. 2016.
\newblock \href {https://doi.org/10.18653/v1/D16-1011} {{Rationalizing Neural
  Predictions}}.
\newblock In \emph{Proceedings of the 2016 Conference on Empirical Methods in
  Natural Language Processing}, pages 107--117. Association for Computational
  Linguistics.

\bibitem[{Leviant and Reichart(2015)}]{leviant2015separated}
Ira Leviant and Roi Reichart. 2015.
\newblock {Separated by an Un-common Language: Towards Judgment Language
  Informed Vector Space Modeling}.
\newblock \emph{arXiv preprint arXiv:1508.00106v5}.

\bibitem[{Li et~al.(2016{\natexlab{a}})Li, Chen, Hovy, and Jurafsky}]{N16-1082}
Jiwei Li, Xinlei Chen, Eduard Hovy, and Dan Jurafsky. 2016{\natexlab{a}}.
\newblock \href {https://doi.org/10.18653/v1/N16-1082} {{Visualizing and
  Understanding Neural Models in NLP}}.
\newblock In \emph{Proceedings of the 2016 Conference of the North American
  Chapter of the Association for Computational Linguistics: Human Language
  Technologies}, pages 681--691. Association for Computational Linguistics.

\bibitem[{Li et~al.(2016{\natexlab{b}})Li, Monroe, and
  Jurafsky}]{li2016understanding}
Jiwei Li, Will Monroe, and Dan Jurafsky. 2016{\natexlab{b}}.
\newblock {Understanding Neural Networks through Representation Erasure}.
\newblock \emph{arXiv preprint arXiv:1612.08220v3}.

\bibitem[{Liang et~al.(2018)Liang, Li, Su, Bian, Li, and Shi}]{liang2017deep}
Bin Liang, Hongcheng Li, Miaoqiang Su, Pan Bian, Xirong Li, and Wenchang Shi.
  2018.
\newblock \href {https://doi.org/10.24963/ijcai.2018/585} {{Deep Text
  Classification Can be Fooled}}.
\newblock In \emph{Proceedings of the Twenty-Seventh International Joint
  Conference on Artificial Intelligence, {IJCAI-18}}, pages 4208--4215.
  International Joint Conferences on Artificial Intelligence Organization.

\bibitem[{Linzen et~al.(2016)Linzen, Dupoux, and
  Goldberg}]{linzen2016assessing}
Tal Linzen, Emmanuel Dupoux, and Yoav Goldberg. 2016.
\newblock \href {http://aclweb.org/anthology/Q16-1037} {{Assessing the Ability
  of LSTMs to Learn Syntax-Sensitive Dependencies}}.
\newblock \emph{Transactions of the Association for Computational Linguistics},
  4:521--535.

\bibitem[{Lipton(2016)}]{lipton2016mythos}
Zachary~C. Lipton. 2016.
\newblock {The Mythos of Model Interpretability}.
\newblock In \emph{ICML Workshop on Human Interpretability of Machine
  Learning}.

\bibitem[{Liu et~al.(2018)Liu, Levy, Schwartz, Tan, and Smith}]{W18-3024}
Nelson~F. Liu, Omer Levy, Roy Schwartz, Chenhao Tan, and Noah~A. Smith. 2018.
\newblock \href {http://aclweb.org/anthology/W18-3024} {{LSTMs Exploit
  Linguistic Attributes of Data}}.
\newblock In \emph{Proceedings of The Third Workshop on Representation Learning
  for NLP}, pages 180--186. Association for Computational Linguistics.

\bibitem[{Liu et~al.(2017)Liu, Chen, Liu, and Song}]{liu2016delving}
Yanpei Liu, Xinyun Chen, Chang Liu, and Dawn Song. 2017.
\newblock {Delving into Transferable Adversarial Examples and Black-box
  Attacks}.
\newblock In \emph{International Conference on Learning Representations
  (ICLR)}.

\bibitem[{Luong et~al.(2013)Luong, Socher, and Manning}]{W13-3512}
Thang Luong, Richard Socher, and Christopher Manning. 2013.
\newblock \href {http://aclweb.org/anthology/W13-3512} {{Better Word
  Representations with Recursive Neural Networks for Morphology}}.
\newblock In \emph{Proceedings of the Seventeenth Conference on Computational
  Natural Language Learning}, pages 104--113. Association for Computational
  Linguistics.

\bibitem[{Maillard and Clark(2018)}]{W18-2903}
Jean Maillard and Stephen Clark. 2018.
\newblock \href {http://aclweb.org/anthology/W18-2903} {{Latent Tree Learning
  with Differentiable Parsers: Shift-Reduce Parsing and Chart Parsing}}.
\newblock In \emph{Proceedings of the Workshop on the Relevance of Linguistic
  Structure in Neural Architectures for NLP}, pages 13--18. Association for
  Computational Linguistics.

\bibitem[{Marelli et~al.(2014)Marelli, Bentivogli, Baroni, Bernardi, Menini,
  and Zamparelli}]{S14-2001}
Marco Marelli, Luisa Bentivogli, Marco Baroni, Raffaella Bernardi, Stefano
  Menini, and Roberto Zamparelli. 2014.
\newblock \href {https://doi.org/10.3115/v1/S14-2001} {{SemEval-2014 Task 1:
  Evaluation of Compositional Distributional Semantic Models on Full Sentences
  through Semantic Relatedness and Textual Entailment}}.
\newblock In \emph{Proceedings of the 8th International Workshop on Semantic
  Evaluation (SemEval 2014)}, pages 1--8. Association for Computational
  Linguistics.

\bibitem[{McCoy et~al.(2018)McCoy, Frank, and Linzen}]{mccoy2018revisiting}
R.~Thomas McCoy, Robert Frank, and Tal Linzen. 2018.
\newblock {Revisiting the poverty of the stimulus: Hierarchical generalization
  without a hierarchical bias in recurrent neural networks}.
\newblock In \emph{Proceedings of the 40th Annual Conference of the Cognitive
  Science Society}.

\bibitem[{Miikkulainen and Dyer(1991)}]{Miikkulainen:1991}
Risto Miikkulainen and Michael~G. Dyer. 1991.
\newblock \href {https://doi.org/10.1207/s15516709cog1503\_2} {{Natural
  Language Processing With Modular Pdp Networks and Distributed Lexicon}}.
\newblock \emph{Cognitive Science}, 15(3):343--399.

\bibitem[{Mikolov et~al.(2010)Mikolov, Karafi{\'a}t, Burget,
  {\v{C}}ernock{\`y}, and Khudanpur}]{mikolov2010recurrent}
Tom{\'a}{\v{s}} Mikolov, Martin Karafi{\'a}t, Luk{\'a}{\v{s}} Burget, Jan
  {\v{C}}ernock{\`y}, and Sanjeev Khudanpur. 2010.
\newblock Recurrent neural network based language model.
\newblock In \emph{Eleventh Annual Conference of the International Speech
  Communication Association}.

\bibitem[{Ming et~al.(2017)Ming, Cao, Zhang, Li, Chen, Song, and
  Qu}]{ming2017understanding}
Yao Ming, Shaozu Cao, Ruixiang Zhang, Zhen Li, Yuanzhe Chen, Yangqiu Song, and
  Huamin Qu. 2017.
\newblock \href {http://www.myaooo.com/projects/rnnvis/} {{Understanding Hidden
  Memories of Recurrent Neural Networks}}.
\newblock In \emph{IEEE Conference on Visual Analytics Science and Technology
  (IEEE VAST 2017)}.

\bibitem[{Montavon et~al.(2018)Montavon, Samek, and M{\"u}ller}]{MONTAVON20181}
Gr{\'e}goire Montavon, Wojciech Samek, and Klaus-Robert M{\"u}ller. 2018.
\newblock \href {https://doi.org/https://doi.org/10.1016/j.dsp.2017.10.011}
  {Methods for interpreting and understanding deep neural networks}.
\newblock \emph{Digital Signal Processing}, 73:1 -- 15.

\bibitem[{Mudrakarta et~al.(2018)Mudrakarta, Taly, Sundararajan, and
  Dhamdhere}]{P18-1176}
Pramod~Kaushik Mudrakarta, Ankur Taly, Mukund Sundararajan, and Kedar
  Dhamdhere. 2018.
\newblock \href {http://aclweb.org/anthology/P18-1176} {{Did the Model
  Understand the Question?}}
\newblock In \emph{Proceedings of the 56th Annual Meeting of the Association
  for Computational Linguistics (Volume 1: Long Papers)}, pages 1896--1906.
  Association for Computational Linguistics.

\bibitem[{Mullenbach et~al.(2018)Mullenbach, Wiegreffe, Duke, Sun, and
  Eisenstein}]{N18-1100}
James Mullenbach, Sarah Wiegreffe, Jon Duke, Jimeng Sun, and Jacob Eisenstein.
  2018.
\newblock \href {http://aclweb.org/anthology/N18-1100} {{Explainable Prediction
  of Medical Codes from Clinical Text}}.
\newblock In \emph{Proceedings of the 2018 Conference of the North American
  Chapter of the Association for Computational Linguistics: Human Language
  Technologies, Volume 1 (Long Papers)}, pages 1101--1111. Association for
  Computational Linguistics.

\bibitem[{Murdoch et~al.(2018)Murdoch, Liu, and Yu}]{james2018beyond}
W.~James Murdoch, Peter~J. Liu, and Bin Yu. 2018.
\newblock \href {https://openreview.net/forum?id=rkRwGg-0Z} {{Beyond Word
  Importance: Contextual Decomposition to Extract Interactions from {LSTM}s}}.
\newblock In \emph{International Conference on Learning Representations}.

\bibitem[{Murphy et~al.(2012)Murphy, Talukdar, and Mitchell}]{C12-1118}
Brian Murphy, Partha Talukdar, and Tom Mitchell. 2012.
\newblock \href {http://www.aclweb.org/anthology/C12-1118} {{Learning Effective
  and Interpretable Semantic Models using Non-Negative Sparse Embedding}}.
\newblock In \emph{Proceedings of COLING 2012}, pages 1933--1950. The COLING
  2012 Organizing Committee.

\bibitem[{Nagamine et~al.(2015)Nagamine, Seltzer, and
  Mesgarani}]{nagamine2015exploring}
Tasha Nagamine, Michael~L. Seltzer, and Nima Mesgarani. 2015.
\newblock {Exploring How Deep Neural Networks Form Phonemic Categories}.
\newblock In \emph{Interspeech 2015}.

\bibitem[{Nagamine et~al.(2016)Nagamine, Seltzer, and Mesgarani}]{Nagamine2016}
Tasha Nagamine, Michael~L. Seltzer, and Nima Mesgarani. 2016.
\newblock \href {https://doi.org/10.21437/Interspeech.2016-1406} {{On the Role
  of Nonlinear Transformations in Deep Neural Network Acoustic Models}}.
\newblock In \emph{Interspeech 2016}, pages 803--807.

\bibitem[{Naik et~al.(2018)Naik, Ravichander, Sadeh, Rose, and
  Neubig}]{naik2018stress}
Aakanksha Naik, Abhilasha Ravichander, Norman Sadeh, Carolyn Rose, and Graham
  Neubig. 2018.
\newblock \href {http://aclweb.org/anthology/C18-1198} {{Stress Test Evaluation
  for Natural Language Inference}}.
\newblock In \emph{Proceedings of the 27th International Conference on
  Computational Linguistics}, pages 2340--2353. Association for Computational
  Linguistics.

\bibitem[{Narodytska and Kasiviswanathan(2017)}]{8014906}
Nina Narodytska and Shiva Kasiviswanathan. 2017.
\newblock \href {https://doi.org/10.1109/CVPRW.2017.172} {{Simple Black-Box
  Adversarial Attacks on Deep Neural Networks}}.
\newblock In \emph{2017 IEEE Conference on Computer Vision and Pattern
  Recognition Workshops (CVPRW)}, pages 1310--1318.

\bibitem[{Niklasson and Lin\r{a}ker(2000)}]{doi:10.1080/09540090010014070}
Lars Niklasson and Fredrik Lin\r{a}ker. 2000.
\newblock \href {https://doi.org/10.1080/09540090010014070} {Distributed
  representations for extended syntactic transformation}.
\newblock \emph{Connection Science}, 12(3-4):299--314.

\bibitem[{Niu and Bansal(2018)}]{Niu:2018}
Tong Niu and Mohit Bansal. 2018.
\newblock \href {http://aclweb.org/anthology/K18-1047} {{Adversarial
  Over-Sensitivity and Over-Stability Strategies for Dialogue Models}}.
\newblock In \emph{Proceedings of the 22nd Conference on Computational Natural
  Language Learning}, pages 486--496. Association for Computational
  Linguistics.

\bibitem[{Papernot et~al.(2016{\natexlab{a}})Papernot, McDaniel, and
  Goodfellow}]{papernot2016transferability}
Nicolas Papernot, Patrick McDaniel, and Ian Goodfellow. 2016{\natexlab{a}}.
\newblock {Transferability in Machine Learning: From Phenomena to Black-Box
  Attacks using Adversarial Samples}.
\newblock \emph{arXiv preprint arXiv:1605.07277v1}.

\bibitem[{Papernot et~al.(2017)Papernot, McDaniel, Goodfellow, Jha, Celik, and
  Swami}]{Papernot:2017:PBA:3052973.3053009}
Nicolas Papernot, Patrick McDaniel, Ian Goodfellow, Somesh Jha, Z.~Berkay
  Celik, and Ananthram Swami. 2017.
\newblock \href {https://doi.org/10.1145/3052973.3053009} {{Practical Black-Box
  Attacks Against Machine Learning}}.
\newblock In \emph{Proceedings of the 2017 ACM on Asia Conference on Computer
  and Communications Security}, ASIA CCS '17, pages 506--519, New York, NY,
  USA. ACM.

\bibitem[{Papernot et~al.(2016{\natexlab{b}})Papernot, McDaniel, Swami, and
  Harang}]{papernot2016crafting}
Nicolas Papernot, Patrick McDaniel, Ananthram Swami, and Richard Harang.
  2016{\natexlab{b}}.
\newblock {Crafting Adversarial Input Sequences for Recurrent Neural Networks}.
\newblock In \emph{Military Communications Conference, MILCOM 2016-2016 IEEE},
  pages 49--54. IEEE.

\bibitem[{Park et~al.(2018)Park, Hendricks, Akata, Rohrbach, Schiele, Darrell,
  and Rohrbach}]{park2018multimodal}
Dong~Huk Park, Lisa~Anne Hendricks, Zeynep Akata, Anna Rohrbach, Bernt Schiele,
  Trevor Darrell, and Marcus Rohrbach. 2018.
\newblock {Multimodal Explanations: Justifying Decisions and Pointing to the
  Evidence}.
\newblock In \emph{The IEEE Conference on Computer Vision and Pattern
  Recognition (CVPR)}.

\bibitem[{Park et~al.(2017)Park, Bak, and Oh}]{D17-1041}
Sungjoon Park, JinYeong Bak, and Alice Oh. 2017.
\newblock \href {http://aclweb.org/anthology/D17-1041} {{Rotated Word Vector
  Representations and their Interpretability}}.
\newblock In \emph{Proceedings of the 2017 Conference on Empirical Methods in
  Natural Language Processing}, pages 401--411. Association for Computational
  Linguistics.

\bibitem[{Peters et~al.(2018)Peters, Neumann, Zettlemoyer, and
  Yih}]{peters2018dissecting}
Matthew Peters, Mark Neumann, Luke Zettlemoyer, and Wen-tau Yih. 2018.
\newblock \href {http://aclweb.org/anthology/D18-1179} {{Dissecting Contextual
  Word Embeddings: Architecture and Representation}}.
\newblock In \emph{Proceedings of the 2018 Conference on Empirical Methods in
  Natural Language Processing}, pages 1499--1509. Association for Computational
  Linguistics.

\bibitem[{Poliak et~al.(2018{\natexlab{a}})Poliak, Haldar, Rudinger, Hu,
  Pavlick, White, and Van~Durme}]{D18-1007}
Adam Poliak, Aparajita Haldar, Rachel Rudinger, J.~Edward Hu, Ellie Pavlick,
  Aaron~Steven White, and Benjamin Van~Durme. 2018{\natexlab{a}}.
\newblock \href {http://aclweb.org/anthology/D18-1007} {{Collecting Diverse
  Natural Language Inference Problems for Sentence Representation Evaluation}}.
\newblock In \emph{Proceedings of the 2018 Conference on Empirical Methods in
  Natural Language Processing}, pages 67--81. Association for Computational
  Linguistics.

\bibitem[{Poliak et~al.(2018{\natexlab{b}})Poliak, Naradowsky, Haldar,
  Rudinger, and Van~Durme}]{S18-2023}
Adam Poliak, Jason Naradowsky, Aparajita Haldar, Rachel Rudinger, and Benjamin
  Van~Durme. 2018{\natexlab{b}}.
\newblock \href {http://aclweb.org/anthology/S18-2023} {{Hypothesis Only
  Baselines in Natural Language Inference}}.
\newblock In \emph{Proceedings of the Seventh Joint Conference on Lexical and
  Computational Semantics}, pages 180--191. Association for Computational
  Linguistics.

\bibitem[{Pollack(1990)}]{POLLACK199077}
Jordan~B. Pollack. 1990.
\newblock \href {https://doi.org/https://doi.org/10.1016/0004-3702(90)90005-K}
  {Recursive distributed representations}.
\newblock \emph{Artificial Intelligence}, 46(1):77 -- 105.

\bibitem[{Qian et~al.(2016{\natexlab{a}})Qian, Qiu, and
  Huang}]{qian-qiu-huang:2016:EMNLP2016}
Peng Qian, Xipeng Qiu, and Xuanjing Huang. 2016{\natexlab{a}}.
\newblock \href {https://aclweb.org/anthology/D16-1079} {{Analyzing Linguistic
  Knowledge in Sequential Model of Sentence}}.
\newblock In \emph{Proceedings of the 2016 Conference on Empirical Methods in
  Natural Language Processing}, pages 826--835, Austin, Texas. Association for
  Computational Linguistics.

\bibitem[{Qian et~al.(2016{\natexlab{b}})Qian, Qiu, and
  Huang}]{qian-qiu-huang:2016:P16-11}
Peng Qian, Xipeng Qiu, and Xuanjing Huang. 2016{\natexlab{b}}.
\newblock \href {http://www.aclweb.org/anthology/P16-1140} {{Investigating
  Language Universal and Specific Properties in Word Embeddings}}.
\newblock In \emph{Proceedings of the 54th Annual Meeting of the Association
  for Computational Linguistics (Volume 1: Long Papers)}, pages 1478--1488,
  Berlin, Germany. Association for Computational Linguistics.

\bibitem[{Ribeiro et~al.(2018)Ribeiro, Singh, and Guestrin}]{P18-1079}
Marco~Tulio Ribeiro, Sameer Singh, and Carlos Guestrin. 2018.
\newblock \href {http://aclweb.org/anthology/P18-1079} {{Semantically
  Equivalent Adversarial Rules for Debugging NLP models}}.
\newblock In \emph{Proceedings of the 56th Annual Meeting of the Association
  for Computational Linguistics (Volume 1: Long Papers)}, pages 856--865.
  Association for Computational Linguistics.

\bibitem[{Rikters(2018)}]{rikters2018debugging}
Mat{\=\i}ss Rikters. 2018.
\newblock \href {http://attention.lielakeda.lv} {{Debugging Neural Machine
  Translations}}.
\newblock \emph{arXiv preprint arXiv:1808.02733v1}.

\bibitem[{Rios~Gonzales et~al.(2017)Rios~Gonzales, Mascarell, and
  Sennrich}]{W17-4702}
Annette Rios~Gonzales, Laura Mascarell, and Rico Sennrich. 2017.
\newblock \href {http://aclweb.org/anthology/W17-4702} {{Improving Word Sense
  Disambiguation in Neural Machine Translation with Sense Embeddings}}.
\newblock In \emph{Proceedings of the Second Conference on Machine
  Translation}, pages 11--19. Association for Computational Linguistics.

\bibitem[{Rockt{\"a}schel et~al.(2016)Rockt{\"a}schel, Grefenstette, Hermann,
  Ko{\v{c}}isk{\`y}, and Blunsom}]{rocktaschel2016reasoning}
Tim Rockt{\"a}schel, Edward Grefenstette, Karl~Moritz Hermann, Tom{\'a}{\v{s}}
  Ko{\v{c}}isk{\`y}, and Phil Blunsom. 2016.
\newblock {Reasoning about Entailment with Neural Attention}.
\newblock In \emph{International Conference on Learning Representations
  (ICLR)}.

\bibitem[{Rozsa et~al.(2016)Rozsa, Rudd, and Boult}]{rozsa2016adversarial}
Andras Rozsa, Ethan~M. Rudd, and Terrance~E. Boult. 2016.
\newblock {Adversarial Diversity and Hard Positive Generation}.
\newblock In \emph{Proceedings of the IEEE Conference on Computer Vision and
  Pattern Recognition Workshops}, pages 25--32.

\bibitem[{Rudinger et~al.(2018)Rudinger, Naradowsky, Leonard, and
  Van~Durme}]{N18-2002}
Rachel Rudinger, Jason Naradowsky, Brian Leonard, and Benjamin Van~Durme. 2018.
\newblock \href {http://aclweb.org/anthology/N18-2002} {{Gender Bias in
  Coreference Resolution}}.
\newblock In \emph{Proceedings of the 2018 Conference of the North American
  Chapter of the Association for Computational Linguistics: Human Language
  Technologies, Volume 2 (Short Papers)}, pages 8--14. Association for
  Computational Linguistics.

\bibitem[{Rumelhart and McClelland(1986)}]{Rumelhart:1986:LPT:21935.42475}
D.~E. Rumelhart and J.~L. McClelland. 1986.
\newblock \href {http://dl.acm.org/citation.cfm?id=21935.42475} {{Parallel
  Distributed Processing: Explorations in the Microstructure of Cognition}}.
\newblock volume~2, chapter {On Learning the Past Tenses of English Verbs},
  pages 216--271. MIT Press, Cambridge, MA, USA.

\bibitem[{Rush et~al.(2015)Rush, Chopra, and Weston}]{D15-1044}
Alexander~M. Rush, Sumit Chopra, and Jason Weston. 2015.
\newblock \href {https://doi.org/10.18653/v1/D15-1044} {{A Neural Attention
  Model for Abstractive Sentence Summarization}}.
\newblock In \emph{Proceedings of the 2015 Conference on Empirical Methods in
  Natural Language Processing}, pages 379--389. Association for Computational
  Linguistics.

\bibitem[{Sakaguchi et~al.(2017)Sakaguchi, Duh, Post, and
  Durme}]{DBLP:conf/aaai/SakaguchiDPD17}
Keisuke Sakaguchi, Kevin Duh, Matt Post, and Benjamin~Van Durme. 2017.
\newblock \href {http://aaai.org/ocs/index.php/AAAI/AAAI17/paper/view/14332}
  {{Robsut Wrod Reocginiton via Semi-Character Recurrent Neural Network}}.
\newblock In \emph{Proceedings of the Thirty-First {AAAI} Conference on
  Artificial Intelligence, February 4-9, 2017, San Francisco, California,
  {USA.}}, pages 3281--3287. {AAAI} Press.

\bibitem[{Samanta and Mehta(2017)}]{samanta2017towards}
Suranjana Samanta and Sameep Mehta. 2017.
\newblock {Towards Crafting Text Adversarial Samples}.
\newblock \emph{arXiv preprint arXiv:1707.02812v1}.

\bibitem[{Sanchez et~al.(2018)Sanchez, Mitchell, and Riedel}]{N18-1179}
Ivan Sanchez, Jeff Mitchell, and Sebastian Riedel. 2018.
\newblock \href {http://aclweb.org/anthology/N18-1179} {{Behavior Analysis of
  NLI Models: Uncovering the Influence of Three Factors on Robustness}}.
\newblock In \emph{Proceedings of the 2018 Conference of the North American
  Chapter of the Association for Computational Linguistics: Human Language
  Technologies, Volume 1 (Long Papers)}, pages 1975--1985. Association for
  Computational Linguistics.

\bibitem[{Sato et~al.(2018)Sato, Suzuki, Shindo, and Matsumoto}]{ijcai2018-601}
Motoki Sato, Jun Suzuki, Hiroyuki Shindo, and Yuji Matsumoto. 2018.
\newblock \href {https://doi.org/10.24963/ijcai.2018/601} {{Interpretable
  Adversarial Perturbation in Input Embedding Space for Text}}.
\newblock In \emph{Proceedings of the Twenty-Seventh International Joint
  Conference on Artificial Intelligence, {IJCAI-18}}, pages 4323--4330.
  International Joint Conferences on Artificial Intelligence Organization.

\bibitem[{Senel et~al.(2018)Senel, Utlu, Yucesoy, Koc, and
  Cukur}]{senel2018semantic}
Lutfi~Kerem Senel, Ihsan Utlu, Veysel Yucesoy, Aykut Koc, and Tolga Cukur.
  2018.
\newblock \href {https://doi.org/10.1109/TASLP.2018.2837384} {{Semantic
  Structure and Interpretability of Word Embeddings}}.
\newblock \emph{IEEE/ACM Transactions on Audio, Speech, and Language
  Processing}.

\bibitem[{Sennrich(2017)}]{E17-2060}
Rico Sennrich. 2017.
\newblock \href {http://aclweb.org/anthology/E17-2060} {{How Grammatical is
  Character-level Neural Machine Translation? Assessing MT Quality with
  Contrastive Translation Pairs}}.
\newblock In \emph{Proceedings of the 15th Conference of the European Chapter
  of the Association for Computational Linguistics: Volume 2, Short Papers},
  pages 376--382. Association for Computational Linguistics.

\bibitem[{Shi et~al.(2018)Shi, Mao, Xiao, Jiang, and Sun}]{shi2018learning}
Haoyue Shi, Jiayuan Mao, Tete Xiao, Yuning Jiang, and Jian Sun. 2018.
\newblock \href {http://aclweb.org/anthology/C18-1315} {{Learning
  Visually-Grounded Semantics from Contrastive Adversarial Samples}}.
\newblock In \emph{Proceedings of the 27th International Conference on
  Computational Linguistics}, pages 3715--3727. Association for Computational
  Linguistics.

\bibitem[{Shi et~al.(2016{\natexlab{a}})Shi, Knight, and Yuret}]{D16-1248}
Xing Shi, Kevin Knight, and Deniz Yuret. 2016{\natexlab{a}}.
\newblock \href {https://doi.org/10.18653/v1/D16-1248} {{Why Neural
  Translations are the Right Length}}.
\newblock In \emph{Proceedings of the 2016 Conference on Empirical Methods in
  Natural Language Processing}, pages 2278--2282. Association for Computational
  Linguistics.

\bibitem[{Shi et~al.(2016{\natexlab{b}})Shi, Padhi, and
  Knight}]{shi-padhi-knight:2016:EMNLP2016}
Xing Shi, Inkit Padhi, and Kevin Knight. 2016{\natexlab{b}}.
\newblock \href {https://aclweb.org/anthology/D16-1159} {{Does String-Based
  Neural MT Learn Source Syntax?}}
\newblock In \emph{Proceedings of the 2016 Conference on Empirical Methods in
  Natural Language Processing}, pages 1526--1534, Austin, Texas. Association
  for Computational Linguistics.

\bibitem[{Singh et~al.(2018)Singh, Murdoch, and Yu}]{singh2018hierarchical}
Chandan Singh, W.~James Murdoch, and Bin Yu. 2018.
\newblock {Hierarchical interpretations for neural network predictions}.
\newblock \emph{arXiv preprint arXiv:1806.05337v1}.

\bibitem[{Strobelt et~al.(2018{\natexlab{a}})Strobelt, Gehrmann, Behrisch,
  Perer, Pfister, and Rush}]{strobelt2018seq2seq}
Hendrik Strobelt, Sebastian Gehrmann, Michael Behrisch, Adam Perer, Hanspeter
  Pfister, and Alexander~M. Rush. 2018{\natexlab{a}}.
\newblock {Seq2Seq-Vis: A Visual Debugging Tool for Sequence-to-Sequence
  Models}.
\newblock \emph{arXiv preprint arXiv:1804.09299v1}.

\bibitem[{Strobelt et~al.(2018{\natexlab{b}})Strobelt, Gehrmann, Pfister, and
  Rush}]{strobelt2018lstmvis}
Hendrik Strobelt, Sebastian Gehrmann, Hanspeter Pfister, and Alexander~M. Rush.
  2018{\natexlab{b}}.
\newblock \href {http://lstm.seas.harvard.edu} {{LSTMVis: A Tool for Visual
  Analysis of Hidden State Dynamics in Recurrent Neural Networks}}.
\newblock \emph{IEEE transactions on visualization and computer graphics},
  24(1):667--676.

\bibitem[{Sundararajan et~al.(2017)Sundararajan, Taly, and
  Yan}]{pmlr-v70-sundararajan17a}
Mukund Sundararajan, Ankur Taly, and Qiqi Yan. 2017.
\newblock \href {http://proceedings.mlr.press/v70/sundararajan17a.html}
  {{Axiomatic Attribution for Deep Networks}}.
\newblock In \emph{Proceedings of the 34th International Conference on Machine
  Learning}, volume~70 of \emph{Proceedings of Machine Learning Research},
  pages 3319--3328, International Convention Centre, Sydney, Australia. PMLR.

\bibitem[{Sutskever et~al.(2014)Sutskever, Vinyals, and
  Le}]{sutskever2014sequence}
Ilya Sutskever, Oriol Vinyals, and Quoc~V. Le. 2014.
\newblock {Sequence to Sequence Learning with Neural Networks}.
\newblock In \emph{Advances in neural information processing systems}, pages
  3104--3112.

\bibitem[{Suzgun et~al.(2019)Suzgun, Belinkov, and Shieber}]{suzgun:2019:SCiL}
Mirac Suzgun, Yonatan Belinkov, and Stuart~M. Shieber. 2019.
\newblock {On Evaluating the Generalization of LSTM Models in Formal
  Languages}.
\newblock In \emph{Proceedings of the Society for Computation in Linguistics
  (SCiL)}.

\bibitem[{Szegedy et~al.(2014)Szegedy, Zaremba, Sutskever, Bruna, Erhan,
  Goodfellow, and Fergus}]{szegedy2013intriguing}
Christian Szegedy, Wojciech Zaremba, Ilya Sutskever, Joan Bruna, Dumitru Erhan,
  Ian Goodfellow, and Rob Fergus. 2014.
\newblock Intriguing properties of neural networks.
\newblock In \emph{International Conference on Learning Representations
  (ICLR)}.

\bibitem[{Tang et~al.(2018)Tang, Sennrich, and Nivre}]{W18-6304}
Gongbo Tang, Rico Sennrich, and Joakim Nivre. 2018.
\newblock \href {http://aclweb.org/anthology/W18-6304} {{An Analysis of
  Attention Mechanisms: The Case of Word Sense Disambiguation in Neural Machine
  Translation}}.
\newblock In \emph{Proceedings of the Third Conference on Machine Translation:
  Research Papers}, pages 26--35. Association for Computational Linguistics.

\bibitem[{Tay et~al.(2018)Tay, Luu, and Hui}]{tay2018couplenet}
Yi~Tay, Anh~Tuan Luu, and Siu~Cheung Hui. 2018.
\newblock {CoupleNet: Paying Attention to Couples with Coupled Attention for
  Relationship Recommendation}.
\newblock In \emph{Proceedings of the Twelfth International AAAI Conference on
  Web and Social Media (ICWSM)}.

\bibitem[{Tran et~al.(2018)Tran, Bisazza, and Monz}]{tran2018importance}
Ke~Tran, Arianna Bisazza, and Christof Monz. 2018.
\newblock \href {http://aclweb.org/anthology/D18-1503} {{The Importance of
  Being Recurrent for Modeling Hierarchical Structure}}.
\newblock In \emph{Proceedings of the 2018 Conference on Empirical Methods in
  Natural Language Processing}, pages 4731--4736. Association for Computational
  Linguistics.

\bibitem[{Vanmassenhove et~al.(2017)Vanmassenhove, Du, and
  Way}]{vanmassenhoveinvestigating}
Eva Vanmassenhove, Jinhua Du, and Andy Way. 2017.
\newblock {Investigating `Aspect' in NMT and SMT: Translating the English
  Simple Past and Present Perfect}.
\newblock \emph{Computational Linguistics in the Netherlands Journal},
  7:109--128.

\bibitem[{Veldhoen et~al.(2016)Veldhoen, Hupkes, and
  Zuidema}]{veldhoen2016diagnostic}
Sara Veldhoen, Dieuwke Hupkes, and Willem Zuidema. 2016.
\newblock {Diagnostic Classifiers: Revealing how Neural Networks Process
  Hierarchical Structure}.
\newblock In \emph{CEUR Workshop Proceedings}.

\bibitem[{Voita et~al.(2018)Voita, Serdyukov, Sennrich, and Titov}]{P18-1117}
Elena Voita, Pavel Serdyukov, Rico Sennrich, and Ivan Titov. 2018.
\newblock \href {http://aclweb.org/anthology/P18-1117} {{Context-Aware Neural
  Machine Translation Learns Anaphora Resolution}}.
\newblock In \emph{Proceedings of the 56th Annual Meeting of the Association
  for Computational Linguistics (Volume 1: Long Papers)}, pages 1264--1274.
  Association for Computational Linguistics.

\bibitem[{Vylomova et~al.(2016)Vylomova, Cohn, He, and
  Haffari}]{vylomova2016word}
Ekaterina Vylomova, Trevor Cohn, Xuanli He, and Gholamreza Haffari. 2016.
\newblock {Word Representation Models for Morphologically Rich Languages in
  Neural Machine Translation}.
\newblock \emph{arXiv preprint arXiv:1606.04217v1}.

\bibitem[{Wang et~al.(2018{\natexlab{a}})Wang, Singh, Michael, Hill, Levy, and
  Bowman}]{wang2018glue}
Alex Wang, Amapreet Singh, Julian Michael, Felix Hill, Omer Levy, and Samuel~R.
  Bowman. 2018{\natexlab{a}}.
\newblock {GLUE: A Multi-Task Benchmark and Analysis Platform for Natural
  Language Understanding}.
\newblock \emph{arXiv preprint arXiv:1804.07461v1}.

\bibitem[{Wang et~al.(2017{\natexlab{a}})Wang, Qian, and Yu}]{Wang2017}
Shuai Wang, Yanmin Qian, and Kai Yu. 2017{\natexlab{a}}.
\newblock \href {https://doi.org/10.21437/Interspeech.2017-1125} {{What Does
  the Speaker Embedding Encode?}}
\newblock In \emph{Interspeech 2017}, pages 1497--1501.

\bibitem[{Wang et~al.(2018{\natexlab{b}})Wang, Pham, Yin, and
  Neubig}]{wang18emnlptrdec}
Xinyi Wang, Hieu Pham, Pengcheng Yin, and Graham Neubig. 2018{\natexlab{b}}.
\newblock \href {https://arxiv.org/abs/1808.09374} {{A Tree-based Decoder for
  Neural Machine Translation}}.
\newblock In \emph{Conference on Empirical Methods in Natural Language
  Processing (EMNLP)}, Brussels, Belgium.

\bibitem[{Wang et~al.(2017{\natexlab{b}})Wang, Chung, and Lee}]{wang2017gate}
Yu-Hsuan Wang, Cheng-Tao Chung, and Hung-yi Lee. 2017{\natexlab{b}}.
\newblock {Gate Activation Signal Analysis for Gated Recurrent Neural Networks
  and Its Correlation with Phoneme Boundaries}.
\newblock In \emph{Interspeech 2017}.

\bibitem[{Weiss et~al.(2018)Weiss, Goldberg, and Yahav}]{P18-2117}
Gail Weiss, Yoav Goldberg, and Eran Yahav. 2018.
\newblock \href {http://aclweb.org/anthology/P18-2117} {{On the Practical
  Computational Power of Finite Precision RNNs for Language Recognition}}.
\newblock In \emph{Proceedings of the 56th Annual Meeting of the Association
  for Computational Linguistics (Volume 2: Short Papers)}, pages 740--745.
  Association for Computational Linguistics.

\bibitem[{Williams et~al.(2018)Williams, Drozdov, and Bowman}]{Q18-1019}
Adina Williams, Andrew Drozdov, and Samuel~R. Bowman. 2018.
\newblock \href {http://aclweb.org/anthology/Q18-1019} {Do latent tree learning
  models identify meaningful structure in sentences?}
\newblock \emph{Transactions of the Association for Computational Linguistics},
  6:253--267.

\bibitem[{Wu and King(2016)}]{wu2016investigating}
Zhizheng Wu and Simon King. 2016.
\newblock Investigating gated recurrent networks for speech synthesis.
\newblock In \emph{2016 IEEE International Conference on Acoustics, Speech and
  Signal Processing (ICASSP)}, pages 5140--5144. IEEE.

\bibitem[{Yang et~al.(2018)Yang, Chen, Hsieh, Wang, and
  Jordan}]{yang2018greedy}
Puyudi Yang, Jianbo Chen, Cho-Jui Hsieh, Jane-Ling Wang, and Michael~I. Jordan.
  2018.
\newblock {Greedy Attack and Gumbel Attack: Generating Adversarial Examples for
  Discrete Data}.
\newblock \emph{arXiv preprint arXiv:1805.12316v1}.

\bibitem[{Yin et~al.(2016)Yin, Sch{\"u}tze, Xiang, and Zhou}]{Q16-1019}
Wenpeng Yin, Hinrich Sch{\"u}tze, Bing Xiang, and Bowen Zhou. 2016.
\newblock \href {http://aclweb.org/anthology/Q16-1019} {{ABCNN: Attention-Based
  Convolutional Neural Network for Modeling Sentence Pairs}}.
\newblock \emph{Transactions of the Association for Computational Linguistics},
  4:259--272.

\bibitem[{Yuan et~al.(2017)Yuan, He, Zhu, and Li}]{yuan2017adversarial}
Xiaoyong Yuan, Pan He, Qile Zhu, and Xiaolin Li. 2017.
\newblock {Adversarial Examples: Attacks and Defenses for Deep Learning}.
\newblock \emph{arXiv preprint arXiv:1712.07107v3}.

\bibitem[{Zaidan et~al.(2007)Zaidan, Eisner, and Piatko}]{N07-1033}
Omar Zaidan, Jason Eisner, and Christine Piatko. 2007.
\newblock \href {http://www.aclweb.org/anthology/N07-1033} {{Using ``Annotator
  Rationales'' to Improve Machine Learning for Text Categorization}}.
\newblock In \emph{Human Language Technologies 2007: The Conference of the
  North American Chapter of the Association for Computational Linguistics;
  Proceedings of the Main Conference}, pages 260--267. Association for
  Computational Linguistics.

\bibitem[{Zhang and Zhu(2018)}]{Zhang2018}
Quan-shi Zhang and Song-chun Zhu. 2018.
\newblock \href {https://doi.org/10.1631/FITEE.1700808} {Visual
  interpretability for deep learning: A survey}.
\newblock \emph{Frontiers of Information Technology {\&} Electronic
  Engineering}, 19(1):27--39.

\bibitem[{Zhang et~al.(2016)Zhang, Marshall, and Wallace}]{D16-1076}
Ye~Zhang, Iain Marshall, and Byron~C. Wallace. 2016.
\newblock \href {https://doi.org/10.18653/v1/D16-1076} {{Rationale-Augmented
  Convolutional Neural Networks for Text Classification}}.
\newblock In \emph{Proceedings of the 2016 Conference on Empirical Methods in
  Natural Language Processing}, pages 795--804. Association for Computational
  Linguistics.

\bibitem[{Zhao et~al.(2018{\natexlab{a}})Zhao, Wang, Yatskar, Ordonez, and
  Chang}]{N18-2003}
Jieyu Zhao, Tianlu Wang, Mark Yatskar, Vicente Ordonez, and Kai-Wei Chang.
  2018{\natexlab{a}}.
\newblock \href {http://aclweb.org/anthology/N18-2003} {{Gender Bias in
  Coreference Resolution: Evaluation and Debiasing Methods}}.
\newblock In \emph{Proceedings of the 2018 Conference of the North American
  Chapter of the Association for Computational Linguistics: Human Language
  Technologies, Volume 2 (Short Papers)}, pages 15--20. Association for
  Computational Linguistics.

\bibitem[{Zhao et~al.(2018{\natexlab{b}})Zhao, Kim, Zhang, Rush, and
  LeCun}]{pmlr-v80-zhao18b}
Junbo Zhao, Yoon Kim, Kelly Zhang, Alexander Rush, and Yann LeCun.
  2018{\natexlab{b}}.
\newblock \href {http://proceedings.mlr.press/v80/zhao18b.html} {{Adversarially
  Regularized Autoencoders}}.
\newblock In \emph{Proceedings of the 35th International Conference on Machine
  Learning}, volume~80 of \emph{Proceedings of Machine Learning Research},
  pages 5902--5911, Stockholmsm{\"a}ssan, Stockholm, Sweden. PMLR.

\bibitem[{Zhao et~al.(2018{\natexlab{c}})Zhao, Dua, and
  Singh}]{zhao2018generating}
Zhengli Zhao, Dheeru Dua, and Sameer Singh. 2018{\natexlab{c}}.
\newblock \href {https://openreview.net/forum?id=H1BLjgZCb} {{Generating
  Natural Adversarial Examples}}.
\newblock In \emph{International Conference on Learning Representations}.

\end{thebibliography}
\bibliographystyle{acl_natbib}




\clearpage

\onecolumn

\appendix

\section*{Supplementary Materials}

\footnotesize
\begin{longtable}{p{3.5cm} p{3cm} p{5.7cm} p{2.3cm}}
\toprule
Reference & Component & Property & Method \\ 
\midrule
\endfirsthead

\multicolumn{4}{r}{Continued from previous page} \\
\toprule
Reference & Component & Property & Method \\ 
\midrule
\endhead

\midrule \caption{A categorization of work trying to find linguistic information in neural networks according to the neural network component investigated, the linguistic property sought, and the analysis method.} \label{tab:ling-info} \\ 
\midrule 
\multicolumn{4}{r}{Continued on next page} \\ 
\endfoot 

 \caption{A categorization of work trying to find linguistic information in neural networks according to the neural network component investigated, the linguistic property sought, and the analysis method.}
\endlastfoot

\cite{elloumi2018analyzing} & \gls{cnn} activations & Style, accent, broadcast program & Classification \\
\cite{belinkov:2017:nips} & \gls{cnn}/\gls{rnn} activations & Phonetic units & Classification \\ 
\cite{dalvi:2019:AAAI} & \gls{nmt} and LM neurons & \gls{pos}, morphology, lexical semantics & Classification \\ 
\cite{D16-1248} & \gls{nmt} encoder neurons & Sentence length & Regression \\ 
\cite{belinkov:2017:ijcnlp} & \gls{nmt} states & \gls{pos}, lexical semantics & Classification \\ 
\cite{belinkov:2017:acl,dalvi:2017:ijcnlp} & \gls{nmt} states & \gls{pos}, morphology & Classification \\ 
\cite{D18-1313} & \gls{nmt} states & Morphology & Classification \\ 
\cite{tran2018importance} & \gls{rnn} / self-attention states & Subject-verb agreement & Likelihood comparison, direct classification \\ 
\cite{wang2017gate} & \gls{rnn} gates & Phoneme boundaries & Change in activation signal \\
\cite{mccoy2018revisiting} & \gls{rnn} sentence embedding & Hierarchical structure & Classification \\ 
\cite{P18-2003} & \gls{rnn} states & \gls{pos}, ancestor label prediction, dependency relation prediction & Classification \\
\cite{shi-padhi-knight:2016:EMNLP2016} & \gls{rnn} states & \gls{pos}, top syntactic sequence, smallest constituent, tense, voice  & Classification \\
\cite{gulordava2018colorless} & \gls{rnn} states & Number agreement & Likelihood comparison \\ 
\cite{linzen2016assessing} & \gls{rnn} states & Subject-verb agreement & Likelihood comparison, direct classification \\ 
\cite{W18-3024} & \gls{rnn} states & Word presence & Direct classification \\ 
\cite{K17-1037} & \gls{rnn} states in audio-visual model & Phonemes, synonyms & Classification, clustering, discrimination   \\ 
\cite{gelderloos-chrupala:2016:COLING} & \gls{rnn} states in language-vision model & Word boundary, word similarity & Classification \\ 
\cite{qian-qiu-huang:2016:EMNLP2016} & \gls{rnn} states/gates & \gls{pos}, syntactic role, gender, case, definiteness, verb form, mood & Classification, correlation \\
\cite{wu2016investigating} & \gls{rnn} states/gates & Acoustic features & Correlation \\
\cite{I17-1004} & Attention weights & \gls{pos}, word alignment & Distribution measures, match with  alignments \\ 
\cite{P18-1117} & Attention weights & Anaphora & Attention score \\ 
\cite{W18-6304} & Attention weights & Word sense disambiguation & Distribution measures \\ 
\cite{Drexler2017AnalysisOA} & Audio-visual \gls{cnn} activations & Phonemes, speakers, word identity & Clustering, discrimination \\ 
\cite{harwath2017learning} & Audio-visual \gls{cnn} embeddings & Word classes & Clustering \\ 
\cite{chrupala2017representations} & Audio-visual \gls{rnn} activations & Utterance length, word presence, homonym disambiguation & Classification, regression, similarity measures \\ 
\cite{peters2018dissecting} & biLM representations (\gls{rnn}, Transformer, gated \gls{cnn})  & \gls{pos}, constituency parsing, coreference & Classification; similarity scores \\ 
\cite{Nagamine2016} & Hidden activations in feed-forward acoustic model & Phonemes, phonetic features & Classification, clustering measures  \\ 
\cite{nagamine2015exploring} & Hidden activations in feed-forward acoustic model & Phonemes, phonetic features, gender & Clustering, average activations by group/label \\ 
\cite{chaabouni2017learning} & Hidden activations in feed-forward audio-visual model  & Phonetic features & Discrimination  \\ 
\cite{vylomova2016word} & NMT word embeddings & synonyms, morphological features & Nearest neighbors \\ 
\cite{N18-1091} & Parser word embeddings & Word features (shape, etc.) & Classification; also other methods \\ 
\cite{W16-2524} & Sentence embeddings & Semantic role, word presence & Classification  \\ 
\cite{adi2017analysis,adi2016fine} & Sentence embeddings & Sentence length, word presence, word order & Classification \\ 
\cite{ahmad2018multi} & Sentence embeddings & Sentence length, word presence, word order; \gls{pos}, word sense disambiguation; sentence order  & Classification \\ 
\cite{Ganesh:2017:IST:3110025.3110083} & Sentence embeddings & Sentence length, word presence, word order; orthography; social tasks & Classification  \\ 
\cite{conneau2018you} & Sentence embeddings & Sentence length, word presence, word order; tree depth, top constituent; main tense, subject/object number, semantic odd man out, coordinate inversion & Classification \\ 
\cite{brunner2018natural} & Sentence embeddings & Synthetic syntactic patterns & Clustering \\ 
\cite{Wang2017} & Speaker embeddings & Speaker, speech content, word order, utterance length, channel, gender, speaking rate  & Classification \\ 
\cite{qian-qiu-huang:2016:P16-11} & Word embeddings & \gls{pos}, dependency relations, morphological features, emotions & Classification \\ 
\cite{kohn:2015:EMNLP} & Word embeddings & \gls{pos}, head \gls{pos}, dependency relation, gender, case, number, tense   & Classification \\ 
\cite{D15-1002} & Word embeddings & Referential attributes & Classification \\ 
\cite{N18-2122} & Word embeddings, vision \gls{cnn}  & Concepts & Similarity measures  \\ 
\bottomrule
\end{longtable}

\begin{table*}[t]
\footnotesize
\centering
\begin{tabular}{p{3.1cm} p{1cm} p{4cm} p{2.7cm} r p{1.8cm} 
}
\toprule
Reference & Task & Phenomena & Languages & Size & Construction 
\\ 
\midrule
\cite{naik2018stress}  &  NLI  &  Antonyms, quantities, spelling, word overlap, negation, length  &  English  & $7596$ &  Automatic  \\ 
\cite{dasgupta2018evaluating}  &  NLI  &  Compositionality  &  English  & $44010$ &  Automatic  \\ 
\cite{N18-1179}  &  NLI  &  Antonyms, hyper/hyponyms  &  English  &  $6279$  &  Semi-auto.  \\ 
\cite{wang2018glue}  &  NLI  &  Diverse semantics  &  English  & $550$ &  Manual  \\ 
\cite{P18-2103}  &  NLI  &  Lexical inference  &  English  & $8193$ &  Semi-auto.  \\ 
\cite{D18-1007} & NLI &  Diverse & English & $570$K & Manual, semi-auto., automatic  \\ 
\cite{W17-4702}  &  MT  &  Word sense disambiguation  &  German$\rightarrow$English/ French  & $13900$ &  Semi-auto.  \\ 
\cite{W17-4705}  &  MT  &  Morphology  &  English$\rightarrow$Czech/Latvian  & $18500$ &  Automatic  \\ 
\cite{E17-2060}  &  MT  &  Polarity, verb-particle constructions, agreement, transliteration  &  English$\rightarrow$German  & $97$K &  Automatic  \\ 
\cite{N18-1118}  &  MT  &  Discourse  &  English$\rightarrow$French  & $400$ &  Manual \\ 
\cite{D17-1263,isabelle2018challenge}  &  MT  &  Morpho-syntax, syntax, lexicon   &  English$\leftrightarrow$French  & $108+506$ &  Manual \\ 
\cite{burchardt2017linguistic}  &  MT  &  Diverse  &  English$\leftrightarrow$German  & $10000$ &  Manual \\  
\cite{linzen2016assessing}  &  LM  &  Subject-verb agreement  &  English  &  $\sim$$1.35$M  &  Automatic  \\
\cite{gulordava2018colorless}  &  LM  &  Number agreement  &  English, Russian, Hebrew, Italian  &  $\sim$$10$K  &  Automatic  \\ 
\cite{N18-2002}  &  Coref.  &  Gender bias  &  English  & $720$ &  Semi-auto.  \\ 
\cite{N18-2003}  &  Coref.  &  Gender bias  &  English  & $3160$ &  Semi-auto.  \\ 
\cite{lake2018} &  \acrshort{seq2seq}  &  Compositionality  &  English  & $20910$ &  Automatic  \\ 
\cite{D18-1277} & \gls{pos} tagging & Noun-verb ambiguity & English & $32654$ & Semi-auto.  \\ 
\bottomrule
\end{tabular}
\caption{A categorization of challenge sets for evaluating neural networks according to the \gls{nlp} task, the linguistic phenomena, the represented languages, the dataset size, and the construction method. }
\label{tab:challenge-sets}
\end{table*}

\begin{table*}[t]
\footnotesize
\centering
\begin{tabular}{l p{1.4cm} p{1.4cm} p{1.2cm} p{6cm}} 
\toprule
Method & Knowledge & Targeted & Unit & Task \\ 
\midrule
\cite{belinkov:2018:ICLR}  &  Black  &  \xmark  &  Char  &  MT  \\ 
\cite{heigold2017robust}  &  Black  &  \xmark  &  Char  & MT, morphology \\
\cite{DBLP:conf/aaai/SakaguchiDPD17}  &  Black  &  \xmark  &  Char  & Spelling correction \\ 
\cite{zhao2018generating} & Black & \cmark, \xmark & Word & MT, \acrlong{nli} \\ 
\cite{gao2018black}  &  Black  &  \xmark &  Char  & Text classification, sentiment  \\ 
\cite{jia-liang:2017:EMNLP2017}  &  Black  &  \xmark  &  Sentence  & Reading comprehension \\ 
\cite{N18-1170}  &  Black  &  \xmark  &  Syntax  & Sentiment, entailment  \\ 
\cite{shi2018learning}  &  Black  &  \xmark  &  Word  &  Image captioning  \\ 
\cite{alzantot2018generating}  &  Black  &  \xmark  &  Word  & \gls{nli}, sentiment \\ 
\cite{kuleshov2018adversarial}  &  Black  &  \xmark  &  Word  & Text classification, sentiment \\ 
\cite{P18-1079}  &  Black  &  \xmark  &  Word  & Reading comprehension, visual QA, sentiment \\ 
\cite{Niu:2018} & Black & \xmark & Word & Dialogue \\ 
\cite{P18-1241}  &  White  &  \cmark  &  Pixels  &  Image captioning  \\ 
\cite{C18-1055}  &  White  &  \cmark  &  Word  &  MT  \\ 
\cite{cheng2018seq2sick}  &  White  &  \cmark  &  Word  &  MT, summarization  \\ 
\cite{P18-1176}  &  White  &  \xmark  &  Word  &  Reading comprehension, visual and table QA \\ 
\cite{papernot2016crafting}  &  White  &  \xmark  &  Word  &  Sentiment \\ 
\cite{samanta2017towards}  &  White  &  \xmark  &  Word  &  Sentiment, gender detection \\ 
\cite{ijcai2018-601}  &  White  &  \xmark  &  Word   & Text classification, sentiment, grammatical error detection \\ 
\cite{liang2017deep}  &  White  &  \cmark  &  Word/Char  &  Text classification \\ 
\cite{P18-2006}  &  White  &  \xmark  &  Word/Char  &  Text classification \\ 
\cite{yang2018greedy}  &  White  &  \xmark  &  Word/Char  &  Text classification \\ 
\bottomrule
\end{tabular}
\caption{A categorization of methods for 
adversarial examples in \gls{nlp}
according to adversary's knowledge (white-box vs.\ black-box), attack specificity (targeted vs.\ non-targeted), the modified linguistic unit (words, characters, etc.), and the attacked task. }
\label{tab:adversarial}
\end{table*}


\end{document}